\documentclass[letterpaper]{article}
\PassOptionsToPackage{table}{xcolor}
\usepackage[preprint]{aaai2027}
\usepackage[hyphens]{url}
\usepackage{graphicx}
\usepackage{natbib}
\usepackage{caption}
\usepackage{algorithm}
\usepackage{algorithmic}
\usepackage{booktabs}
\usepackage{array}
\usepackage{multirow}
\usepackage{tabularx}
\usepackage{subcaption}
\usepackage{amsmath}
\usepackage{amssymb}
\usepackage{mathtools}
\usepackage{amsthm}

\newtheorem{proposition}{Proposition}
\newtheorem{corollary}[proposition]{Corollary}
\newcolumntype{Y}{>{\centering\arraybackslash}X}
\definecolor{TAIAccent}{RGB}{0,118,118}
\definecolor{OursBlue}{HTML}{EEF5FF}
\newcommand{\gain}[1]{\textcolor{TAIAccent}{#1}}
\newcommand{\best}[1]{\textbf{#1}}
\newcommand{\ours}[1]{\textbf{#1}}

\title{Stepwise Intrinsic Rewards for Reasoning in Large Language Models}
\author{
\large
Xiangwei Wang\textsuperscript{1},
Wei Wang\textsuperscript{1},
Ken Chen\textsuperscript{1},
Nanduni Nimalsiri\textsuperscript{2},
Sachith Seneviratne\textsuperscript{1},
Saman Halgamuge\textsuperscript{1}
}
\affiliations{
\textsuperscript{1}The University of Melbourne, Melbourne, Australia\\
\textsuperscript{2}CSIRO, Australia
}

\begin{document}

\maketitle

\begin{abstract}
Reinforcement learning (RL) has become a widely used paradigm for improving the
reasoning abilities of large language models (LLMs) and Vision-language models
(VLMs). Sparse binary outcome rewards, however, score only final correctness
and cannot identify which intermediate steps contributed to it; in multimodal
tasks, they may also reward answers driven by linguistic priors rather than
visual evidence. Process reward models (PRMs) densify supervision but usually
require process annotations, auxiliary models, or inference-time search. In this paper, we
introduce Stepwise Marginal Information Gain (MIG), an intrinsic process reward
computed from the policy itself. MIG measures how each structured reasoning
prefix changes the length-normalized, teacher-forced log-likelihood of the
reference answer. A monotonic historical watermark rewards only new likelihood
maxima, avoiding duplicate credit after sub-record detours. We combine this
signal with outcome and format rewards and a gated self-distillation objective
that retains only structurally valid and correct trajectories. For VLMs, a
real-versus-blank likelihood gate down-weights rewards when answers remain
predictable without the image. Across eight task-specific benchmarks, the full
method exceeds outcome-only GRPO in every single-run comparison. In broad-data
transfer, it improves average accuracy by up to 4.8 points over binary-reward
training and gains 12.6 points on MathVerse. At 7B, it exceeds an external
PRM-BoN@16 baseline by 12.9 points on vision-language transfer without
inference-time reranking. These results support policy-derived stepwise credit
as an annotation-free alternative to explicit process reward modeling.
\end{abstract}

\section{Introduction}

Reinforcement learning with verifiable rewards (RLVR) has become a standard
way to improve reasoning in large language models (LLMs)
\cite{deepseek-ai_deepseek-r1_2025,yue_does_2025}. Its appeal is simple: a
rule-based evaluator can score the final answer without requiring annotated
reasoning traces.

However, a binary terminal reward assigns identical credit to the whole
trajectory, obscuring decisive steps, redundant continuation, and the source
of near-correct failures.

\newpage
\begin{center}
\begin{minipage}{\columnwidth}
    \centering
    \includegraphics[
        width=\columnwidth,
        height=0.43\columnwidth
    ]{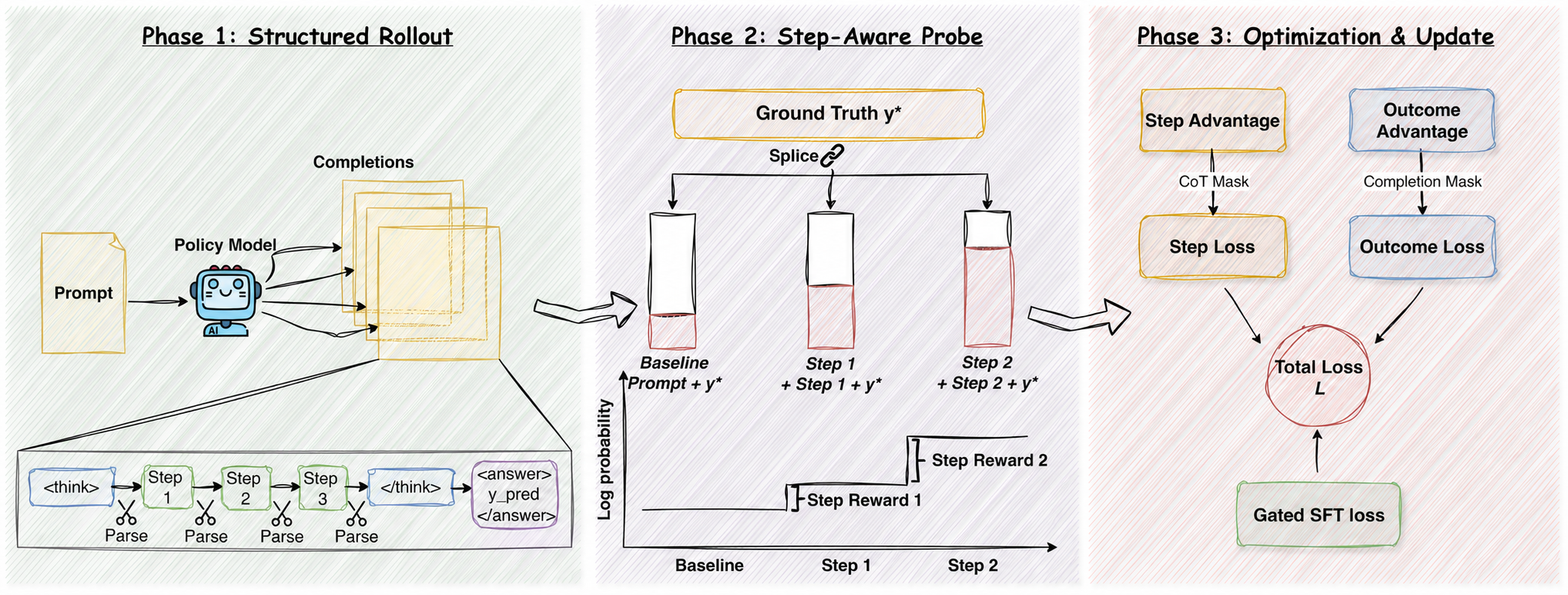}
    \captionof{figure}{Stepwise Marginal Information Gain (MIG) pipeline:
    structured rollout, step-aware probing, and decoupled process, outcome,
    and gated-SFT optimization.}
    \label{fig:method_overview}
\end{minipage}
\end{center}

Process reward models (PRMs) address this limitation by evaluating
intermediate steps \cite{zhang_lessons_2025,zhang_linking_2025}, but typically
introduce process annotations, Monte Carlo labels, an auxiliary model, or
best-of-$N$ search. These costs are especially restrictive when moving across
domains and modalities.

Verifier-free methods instead use the policy's probability of a reference
answer \cite{zhou_reinforcing_2025,tang_beyond_2025}, but a trajectory-level
score still does not identify which step produced the improvement.
Vision-language models introduce an additional ambiguity: a correct answer can
arise from language priors even when the image is ignored.

These limitations impose four requirements on a useful intrinsic process
signal. It should be content-aware rather than tied to step position,
conservative under likelihood drops and recoveries, separable from terminal
correctness, and, for multimodal problems, sensitive to whether the visual
input changes the answer distribution. Satisfying only one requirement is
insufficient. For example, final-answer likelihood supplies a semantic target
but not temporal credit, while a step reward combined directly with
correctness cannot reveal whether a visual model answered from the question
alone.

We propose \textbf{Stepwise Marginal Information Gain (MIG)}, an intrinsic
process reward computed from the policy model. After each structured reasoning
step, MIG measures the change in the teacher-forced likelihood of the reference
answer. A \textbf{monotonic historical watermark} grants positive reward only
when a prefix exceeds every earlier prefix. The resulting reward is
content-aware and position-independent: a useful transformation is credited
wherever it occurs, while likelihood recovery after a sub-record detour
receives no duplicate credit.
Figure~\ref{fig:method_overview} summarizes the rollout, probing, and
optimization phases.

MIG is not itself a correctness reward. We therefore combine it with separate
outcome and format advantages and apply supervised self-distillation only to
trajectories that are both structurally valid and correct. For multimodal
tasks, a visual gate compares answer likelihood under the real image and a
blank counterfactual, softly reducing rewards when the same answer remains
predictable without visual evidence. This design encourages visual dependence
while retaining outcome grounding.

Our contributions are:
\begin{itemize}
    \item We introduce an annotation-free, stepwise process reward based on
    record-setting improvements in reference-answer likelihood.
    \item We show that HWM-MIG telescopes to the best-so-far likelihood
    improvement while assigning positive credit exactly to record-setting
    steps, and combine it with decoupled outcome and gated-SFT objectives.
    \item We extend the reward to vision-language reasoning through a
    counterfactual visual gate and evaluate text and multimodal transfer at 3B
    and 7B scales.
\end{itemize}

\section{Related Work}

\paragraph{Outcome-supervised reasoning.}
Group Relative Policy Optimization (GRPO) adapts PPO-style updates
\cite{schulman_ppo_2017} while avoiding a learned value model by normalizing
rewards among completions sampled for the same prompt
\cite{shao_deepseekmath_2024}. DAPO, GSPO, and GDPO subsequently refine
clipping, sequence-level importance ratios, and multi-reward normalization
\cite{yu_dapo_2025,zheng_group_2025,liu_gdpo_2026}. These advances improve the
policy optimizer, but a terminal score still assigns no temporal structure
within a completion: all reasoning tokens inherit the result of the final
answer. Our focus is complementary. We retain group-relative optimization but
construct an intrinsic process channel that identifies where answer support
first increases.

\paragraph{Process and likelihood supervision.}
PRMs score intermediate states using human judgments, rollout-estimated labels,
or a separately trained verifier
\cite{uesato_process_outcome_2022,lightman_verify_2024,wang_math_shepherd_2024,zhang_r1-vl_2025,zhang_lessons_2025}.
Conditional reward modeling further links process scores to terminal outcomes
\cite{zhang_linking_2025}. Such models can support search and reranking, but
their annotation, training, and inference costs must be repeated across
domains. Annotation-light alternatives learn implicit process rewards from
outcomes or estimate intermediate values with Monte Carlo continuations
\cite{cui_prime_2026,kazemnejad_vineppo_2025}; these still maintain a learned
scorer or sample continuations from intermediate states. Verifier-free methods
instead optimize the policy's probability of a reference answer
\cite{zhou_reinforcing_2025,tang_beyond_2025}, and latent
reasoning objectives use the same model to generate and assess candidate
reasoning \cite{chen_language_2024}. A single likelihood after the full trace,
however, remains a trajectory-level reward. MIG probes every structured prefix
and assigns the resulting improvement to record-setting steps, providing
temporal resolution without an auxiliary scorer.

MIG is also related to delayed-reward redistribution. RUDDER learns a
contribution decomposition of an observed return to produce a return-equivalent
decision process \cite{arjona_medina_rudder_2019}. MIG instead defines its
process return directly from reference-answer likelihood, with no learned
decomposition model and no claim that it preserves the optimal policy of the
original terminal-reward problem.

\paragraph{Self-training and multimodal grounding.}
STaR and ReST improve models by filtering self-generated trajectories using
answer quality \cite{zelikman_star_2022,gulcehre_reinforced_2023}. Our
gated-SFT term follows this outcome-grounded principle, but is optimized jointly
with separately normalized process and outcome advantages. Multimodal RL
systems have also shown that outcome rewards can improve visual reasoning
\cite{zhang_r1-vl_2025,chen_g1_2025,meng2025mmeureka}. Correctness alone does
not establish that the image was used. Counterfactual VQA methods expose
language bias by intervening on visual or linguistic evidence
\cite{niu_counterfactual_2021}. Our gate similarly asks whether the same
reference answer remains predictable after replacing the image, but uses that
diagnostic to scale RL rewards rather than to alter the inference rule.

\section{Method}
\label{sec:method}

\subsection{Structured Rollouts and Step Likelihood}

For a prompt $x$, policy $\pi_\theta$ samples a group of $G$ completions. Each
completion contains a reasoning chain $z$ and final answer $y$. We require
explicit \texttt{\#\#\# Step} markers inside the reasoning span and parse
$z$ into $K$ steps $(s_1,\ldots,s_K)$. Malformed trajectories receive zero
process and format rewards. Robust marker matching is detailed in the
supplementary material.

The format reward is deliberately stricter than detecting a marker substring.
It equals one only for unique, non-nested outer spans: reasoning is enclosed by
\texttt{<think>} and \texttt{</think>}, followed by an answer enclosed by
\texttt{<answer>} and \texttt{</answer>}. No text may occur outside those
spans, and the reasoning body must decompose into a valid sequence of
non-trivial steps. The same parser determines both the format reward and the
token spans used for process optimization, so malformed boundaries cannot
receive one reward channel while being interpreted differently by another.

Let $\mathcal{Y}^*$ contain one or more semantically equivalent reference
answers. For prefix $c_k=(x,s_{1:k})$, define the length-normalized,
teacher-forced answer log-likelihood
\begin{equation}
\ell_k =
\max_{y^*\in\mathcal{Y}^*}
\frac{1}{|y^*|}
\sum_{t=1}^{|y^*|}
\log \pi_\theta(y_t^*\mid c_k,y_{<t}^*),
\label{eq:step_likelihood}
\end{equation}
where $c_0=x$. The task-specific main experiments use the canonical answer;
the supplementary material reports a separate answer-variant study.

Teacher forcing makes the scoring target identical across the $G$ sampled
reasoning chains: only the conditioning prefix changes. Length normalization
prevents longer answer strings from being penalized solely for containing more
tokens, and the maximum over $\mathcal Y^*$ treats approved answer variants as
alternative realizations rather than independent rewards. The generated answer
$y$ is not substituted for $y^*$ in Equation~\ref{eq:step_likelihood}; doing so
would measure self-consistency with the model's own prediction rather than
progress toward the verified target.

\paragraph{Chunked reward computation.}
For a rollout batch with $B$ prompts, $G$ generations per prompt, and
$\bar K$ parsed steps per generation, MIG requires approximately
$BG\bar K$ step-conditioned answer-likelihood evaluations. We flatten these
trajectory-prefix pairs and evaluate them in chunks of 16 for text models and
8 for vision-language models. Every prefix is still evaluated exactly, so
chunking changes peak memory and throughput but not $\ell_k$ or the resulting
MIG rewards. Ignoring padding, the added teacher-forced scoring work scales as
$O(BG\bar K |y^*|)$ answer tokens. It is incurred only during training: MIG
introduces no reward-model parameters, candidate reranking, or additional
forward pass at deployment.

\subsection{Marginal Information Gain}
\label{sec:mig}

Raw likelihood differences can reward a trajectory for first reducing answer
confidence and then recovering it. We prevent this ``pump-and-dump'' pattern
with a historical watermark:
\begin{equation}
h_0=\ell_0,\quad
h_k=\max(h_{k-1},\ell_k),\quad
g_k=(\ell_k-h_{k-1})_+ .
\label{eq:mig}
\end{equation}
Thus, $g_k$ is positive only when step $k$ makes the reference answer more
predictable than after every previous prefix. The trajectory-level process
score is the unnormalized sum
$R_{\mathrm{MIG}}=\sum_{k=1}^{K}g_k$. We do not divide this score by $K$;
the raw sum is the quantity used by the objective and by the telescoping
identity below.

\begin{proposition}[Record-Improvement Potential Difference]
\label{prop:hwm_record}
For scores $\ell_{0:K}$, define $h_0=\ell_0$,
$h_k=\max(h_{k-1},\ell_k)$, and $g_k=(\ell_k-h_{k-1})_+$. Then
\[
    h_k=\max_{0\le j\le k}\ell_j,\qquad g_k=h_k-h_{k-1},
\]
and therefore
\[
    R_{\mathrm{MIG}}=\sum_{k=1}^{K}g_k
    =\max_{0\le j\le K}\ell_j-\ell_0 .
\]
Moreover, $g_k>0$ iff $\ell_k>\max_{0\le j<k}\ell_j$. With
$\tilde s_k=(x,s_{1:k},h_k)$ and $\Phi(\tilde s_k)=h_k$,
\[
    g_k=\Phi(\tilde s_k)-\Phi(\tilde s_{k-1}).
\]
Thus, MIG is a potential difference on the watermark-augmented state.
\end{proposition}

\begin{corollary}[Invariance to Sub-Record Detours]
\label{cor:subrecord_detour}
Inserting after any $m<K$ a finite detour whose scores are all at most $h_m$,
while leaving subsequent scores unchanged, gives every inserted step zero
reward and leaves $R_{\mathrm{MIG}}$ unchanged.
\end{corollary}

\subsection{Decoupled Hybrid Optimization}
\label{sec:hybrid}

Within each sampled group, we separately normalize the cumulative process
score, format reward, and exact-answer reward:
\begin{align}
A_i^{\mathrm{step}} &=
\operatorname{GroupNorm}(R_{\mathrm{MIG}}^{(i)}),\\
A_i^{\mathrm{out}} &=
\operatorname{GroupNorm}(r_{\mathrm{fmt}}^{(i)})
+\operatorname{GroupNorm}(r_{\mathrm{acc}}^{(i)}).
\label{eq:advantages}
\end{align}
For any reward channel $q$, we use
\begin{equation}
\operatorname{GroupNorm}(q_i)
=\frac{q_i-\mu_q}{\sigma_q+\epsilon},
\label{eq:group_norm}
\end{equation}
where the statistics are computed only among the $G$ completions generated for
the same prompt. Channel-wise normalization is performed before the weighted
losses are combined. This preserves the within-group ordering of each signal
without allowing the numerical scale of likelihood gains to absorb the binary
format or accuracy channels.

The process advantage is applied only to reasoning tokens, whereas the outcome
advantage is applied to the full completion. This avoids conflating
intermediate progress with terminal correctness. In particular, a completion
can obtain positive process credit yet still receive a negative outcome
advantage when its emitted answer is wrong; conversely, a correct answer does
not imply that every reasoning prefix improved answer support.

Let $M_{\mathrm{think}}^{(i)}$ select tokens inside the reasoning span and
$M_{\mathrm{comp}}^{(i)}$ select the complete generated response. Writing
$\mathcal L_{\mathrm{GRPO}}(A;M)$ for the standard clipped token-level GRPO
surrogate restricted to mask $M$, the two policy terms are
\begin{equation}
\begin{aligned}
\mathcal L_{\mathrm{MIG}}
&=\mathcal L_{\mathrm{GRPO}}
  (A^{\mathrm{step}};M_{\mathrm{think}}),\\
\mathcal L_{\mathrm{out}}
&=\mathcal L_{\mathrm{GRPO}}
  (A^{\mathrm{out}};M_{\mathrm{comp}}).
\end{aligned}
\label{eq:decoupled_masks}
\end{equation}
Thus, decoupling changes both the reward channel and the tokens to which its
advantage is assigned. The likelihood scores used to construct rewards are
detached during optimization; gradients enter through the GRPO log
probabilities and the self-distillation term, not through
Equation~\ref{eq:step_likelihood}. This avoids turning the reward-scoring pass
into a second, implicit supervised objective.

We additionally distill a sampled completion only when it has valid structure
and a correct final answer:
\begin{equation}
\mathcal{L}_{\mathrm{GSFT}}
=-\frac{1}{G}\sum_i
\omega_{\mathrm{fmt}}^{(i)}\omega_{\mathrm{acc}}^{(i)}
\log\pi_\theta(z^{(i)},y^{(i)}\mid x).
\label{eq:gated_sft}
\end{equation}
The total objective is
\begin{equation}
\mathcal{L}=
\lambda_{\mathrm{out}}\mathcal{L}_{\mathrm{out}}
+\lambda_{\mathrm{step}}\mathcal{L}_{\mathrm{MIG}}
+\lambda_{\mathrm{sft}}\mathcal{L}_{\mathrm{GSFT}},
\label{eq:total_loss}
\end{equation}
where the first two terms use the standard GRPO policy objective with the
advantages and token masks above.

\begin{algorithm}[t]
\caption{Stepwise MIG Policy Optimization}
\label{alg:mig_training}
\begin{algorithmic}[1]
\STATE \textbf{Input:} dataset $\mathcal D$, policy $\pi_\theta$, group size $G$
\REPEAT
\STATE Sample prompts and $G$ completions per prompt
\STATE Parse each completion into $s_{1:K}$ and answer $y$
\FOR{each parsed completion $i$}
    \STATE Compute $\ell_0$ from Equation~\ref{eq:step_likelihood}
    \STATE Set $h_0\leftarrow\ell_0$, $R_{\mathrm{MIG}}^{(i)}\leftarrow0$
    \FOR{$k=1,\ldots,K$}
        \STATE Compute $\ell_k$; set $g_k\leftarrow(\ell_k-h_{k-1})_+$
        \STATE $h_k\leftarrow\max(h_{k-1},\ell_k)$
        \STATE $R_{\mathrm{MIG}}^{(i)}\leftarrow R_{\mathrm{MIG}}^{(i)}+g_k$
    \ENDFOR
    \STATE Compute $r_{\mathrm{fmt}}^{(i)}$ and $r_{\mathrm{acc}}^{(i)}$
    \STATE Set SFT gates
    $\omega_{\mathrm{fmt}}^{(i)}\leftarrow r_{\mathrm{fmt}}^{(i)}$ and
    $\omega_{\mathrm{acc}}^{(i)}\leftarrow r_{\mathrm{acc}}^{(i)}$
    \IF{the input contains an image}
        \STATE Compute $S_{\mathrm{vis}}^{(i)}$ from Equation~\ref{eq:visual_gate}
        \STATE Scale $R_{\mathrm{MIG}}^{(i)}$ and $r_{\mathrm{acc}}^{(i)}$ by
        $S_{\mathrm{vis}}^{(i)}$
    \ENDIF
\ENDFOR
\STATE Group-normalize each reward channel as in Equation~\ref{eq:advantages}
\STATE Compute Equation~\ref{eq:total_loss} with decoupled token masks
\STATE Update $\theta$ by gradient descent
\UNTIL{training completes}
\end{algorithmic}
\end{algorithm}

\subsection{Counterfactual Visual Gate}
\label{sec:visual_gate}

A multimodal model can receive a correct-answer reward while relying primarily
on the question. For a generated reasoning chain $z$, we therefore score the
same reference answer under the real image $I$ and a blank counterfactual
$I_{\varnothing}$:
\begin{align}
\ell_{\mathrm{real}} &=
\operatorname{LL}_\theta(y^*\mid x_{\mathrm{text}},I,z),\\
\ell_{\mathrm{blank}} &=
\operatorname{LL}_\theta(y^*\mid x_{\mathrm{text}},I_{\varnothing},z),\\
S_{\mathrm{vis}} &=
\sigma\!\left(\tau[
\ell_{\mathrm{real}}-\ell_{\mathrm{blank}}-m]\right).
\label{eq:visual_gate}
\end{align}
The chain and text are fixed, so the gap isolates sensitivity to the image
intervention. We use $\tau=5$ and $m=0.1$. The gate scales the process and
accuracy channels,
\begin{equation}
\widetilde R_{\mathrm{MIG}}=S_{\mathrm{vis}}R_{\mathrm{MIG}},
\qquad
\widetilde r_{\mathrm{acc}}=S_{\mathrm{vis}}r_{\mathrm{acc}},
\end{equation}
before group normalization. It does not gate the format reward or
$\mathcal L_{\mathrm{GSFT}}$. Hence the mechanism encourages visual
sensitivity but does not replace correctness grounding.

Applying the gate before group normalization is deliberate: it changes the
relative reward of real-image-dependent trajectories within each sampled
group. Normalizing the visual gap as a separate additive channel would instead
discard a group-constant shift and would not directly condition the process and
accuracy rewards on visual dependence.

\paragraph{Scope of the intervention.}
The gate is global to a sampled trajectory: the same $S_{\mathrm{vis}}$ scales
all of its MIG increments and its binary accuracy reward. We do not subtract a
counterfactual score independently at every step, which would entangle
step-level credit with variation introduced by the counterfactual input. The
blank image is deliberately simple and model-agnostic, but it is only a proxy
for visual dependence. We therefore evaluate the gate both by downstream
accuracy and by controlled answer-changing image edits.

\section{Experiments}
\label{sec:experiments}

The evaluation addresses three questions: whether dense intrinsic credit
improves outcome-only GRPO in task-specific training; whether the gain transfers
when training and evaluation distributions differ; and which components of the
hybrid objective account for the result. We test the first question with eight
complete training curves and in-domain evaluations, the second at both 3B and
7B scales, and the third through reward, SFT, and visual-gate ablations in the
broad-data protocol.

\subsection{Setup}

We use two protocols. In \textbf{task-specific training}, a 3B model is trained
on each target dataset and evaluated on its held-out test set. Text tasks are
GSM8K and MATH \cite{cobbe_training_2021,hendrycks_measuring_2021}, plus the
Chinese and English Tal-SCQ5K science sets \cite{liu_matheval_2025}. Vision
tasks are CMM-Math \cite{liu_cmm-math_2025}, ChartQA
\cite{masry_chartqa_2022}, Super-CLEVR \cite{li_super-clevr_2023}, and
CLEVR-CoGenT \cite{johnson_clevr_2017,chen_g1_2025}.

In \textbf{broad-data transfer}, text-only Qwen2.5-Instruct models
\cite{qwen25_2024} are trained on WebData \cite{zhou_reinforcing_2025} and
evaluated on MMLU-Pro, GSM8K,
MATH-500, and CommonsenseQA
\cite{wang2024mmlupro,talmor2019commonsenseqa}. Qwen2.5-VL-Instruct models are
trained on MMK12 \cite{bai_qwen25vl_2025,meng2025mmeureka} and evaluated on
MathVista, MMStar, ChartQA, and MathVerse
\cite{lu2023mathvista,chen2024we,zhang2024mathverse}. We evaluate both 3B and
7B scales. Pass@1 accuracy is reported from single runs; comparisons are
therefore descriptive rather than statistical significance claims.

\paragraph{Compared methods.}
\textit{Base} is the pretrained instruction model without RL.
\textit{GRPO} in the task-specific study and \textit{Binary reward} in the
transfer study are outcome-only baselines trained with exact-answer and format
rewards. \textit{CoT-likelihood} replaces step decomposition with the
teacher-forced reference-answer likelihood after the complete generated
reasoning chain, $\ell_K$. \textit{MIG only} retains the HWM process signal and
the format reward, but removes the binary outcome and gated-SFT channels.
\textit{Redist-Prox} uses the
same trajectory-level HWM mass but routes it across reasoning steps using
likelihood-proximity weights. The \textit{w/o SFT} and \textit{w/o visual gate}
rows remove one component from the full objective. Finally,
\textit{PRM-BoN@16} uses an external process reward model to select one answer
from 16 candidates; all other rows evaluate one sampled completion and incur
no inference-time reranking. These comparisons separate three issues that a
single binary baseline cannot: whether answer likelihood is useful, whether it
must be localized over prefixes, and whether process credit remains dependent
on outcome and visual grounding.

\paragraph{Metric and aggregation.}
We use exact-answer Pass@1 accuracy for task-specific evaluation and the
standard benchmark answer matcher for broad-data evaluation. Each reported
``Avg.'' in Table~\ref{tab:broad_transfer} is the unweighted arithmetic mean of
the four benchmark accuracies in that block. No answer variants are used in
the task-specific results in Table~\ref{tab:indomain_all}; the separate
variant-aware MATH study is reported in the supplementary material.

All methods in a comparison use the same model, data, and hardware protocol.
We sample $G=8$ trajectories, with a maximum prompt length of 1024 and a
maximum completion length of 2048 at 3B and 4096 at 7B. Rollout and evaluation
temperatures are 1.0 and 0.6. We set
$(\lambda_{\mathrm{out}},\lambda_{\mathrm{step}},\lambda_{\mathrm{sft}})
=(1,0.01,0.001)$. We selected these coefficients by monitoring the gradient
magnitudes induced by the three objective terms during training and balancing
their effective update strengths. Likelihood prefixes are evaluated in chunks
to bound memory.
The 3B runs use four NVIDIA A100 GPUs for training and one A100 for evaluation;
the 7B runs use four NVIDIA H200 GPUs for training and one H200 for evaluation.
All experiments use PyTorch and DeepSpeed ZeRO-3
\cite{rajbhandari_zero_2020}. Full prompts, optimizer settings, chunk sizes,
and hardware details are in the supplementary material.

\subsection{Task-Specific Results}

Table~\ref{tab:indomain_all} reports the 3B in-domain evaluation. MIG improves
over GRPO on all eight datasets. The text gains range from 2.2 points on GSM8K
to 11.0 points on both Tal-SCQ5K variants. Vision-language gains are largest
on Super-CLEVR and CLEVR-CoGenT (8.0 and 9.0 points), and remain positive on
ChartQA and CMM-Math. The pattern is not limited to tasks on which outcome-only
RL already improves the base model. On Tal-SCQ5K-EN, for example, GRPO is two
points below Base whereas Ours is nine points above Base. On the synthetic
visual tasks both RL methods improve strongly, but the additional MIG gain
remains eight to nine points. These are single-run comparisons, so the table
supports consistency across the evaluated tasks rather than a variance or
significance claim.

\begin{table*}[t]
\centering
\small
\setlength{\tabcolsep}{4.2pt}
\begin{tabular*}{\textwidth}{@{\extracolsep{\fill}}lllrrrr@{}}
\toprule
\textbf{Modality} & \textbf{Dataset} & \textbf{Type} &
\textbf{Base} & \textbf{GRPO} & \textbf{Ours} & \textbf{$\Delta$} \\
\midrule
\multicolumn{7}{@{}l}{\textit{Text reasoning}} \\
Text & GSM8K & Math & 78.8 & 81.0 & \best{83.2} & \gain{+2.2} \\
Text & MATH & Math & 56.0 & 58.0 & \best{60.5} & \gain{+2.5} \\
Text & Tal-SCQ5K-CN & Science & 49.0 & 53.0 & \best{64.0} & \gain{+11.0} \\
Text & Tal-SCQ5K-EN & Science & 64.0 & 62.0 & \best{73.0} & \gain{+11.0} \\
\midrule
\multicolumn{7}{@{}l}{\textit{Vision-language reasoning}} \\
Vision & Super-CLEVR & 3D logic & 46.0 & 89.0 & \best{97.0} & \gain{+8.0} \\
Vision & CLEVR-CoGenT & Composition & 54.0 & 82.0 & \best{91.0} & \gain{+9.0} \\
Vision & ChartQA & Charts & 79.0 & 81.0 & \best{87.0} & \gain{+6.0} \\
Vision & CMM-Math & Math & 15.0 & 30.0 & \best{36.0} & \gain{+6.0} \\
\bottomrule
\end{tabular*}
\caption{Task-specific in-domain Pass@1 accuracy (\%) with 3B models.
$\Delta$ is the absolute improvement of Ours over GRPO.}
\label{tab:indomain_all}
\end{table*}

\begin{figure*}[t]
\centering
\begin{minipage}[t]{0.235\textwidth}
    \centering
    \includegraphics[width=\linewidth]{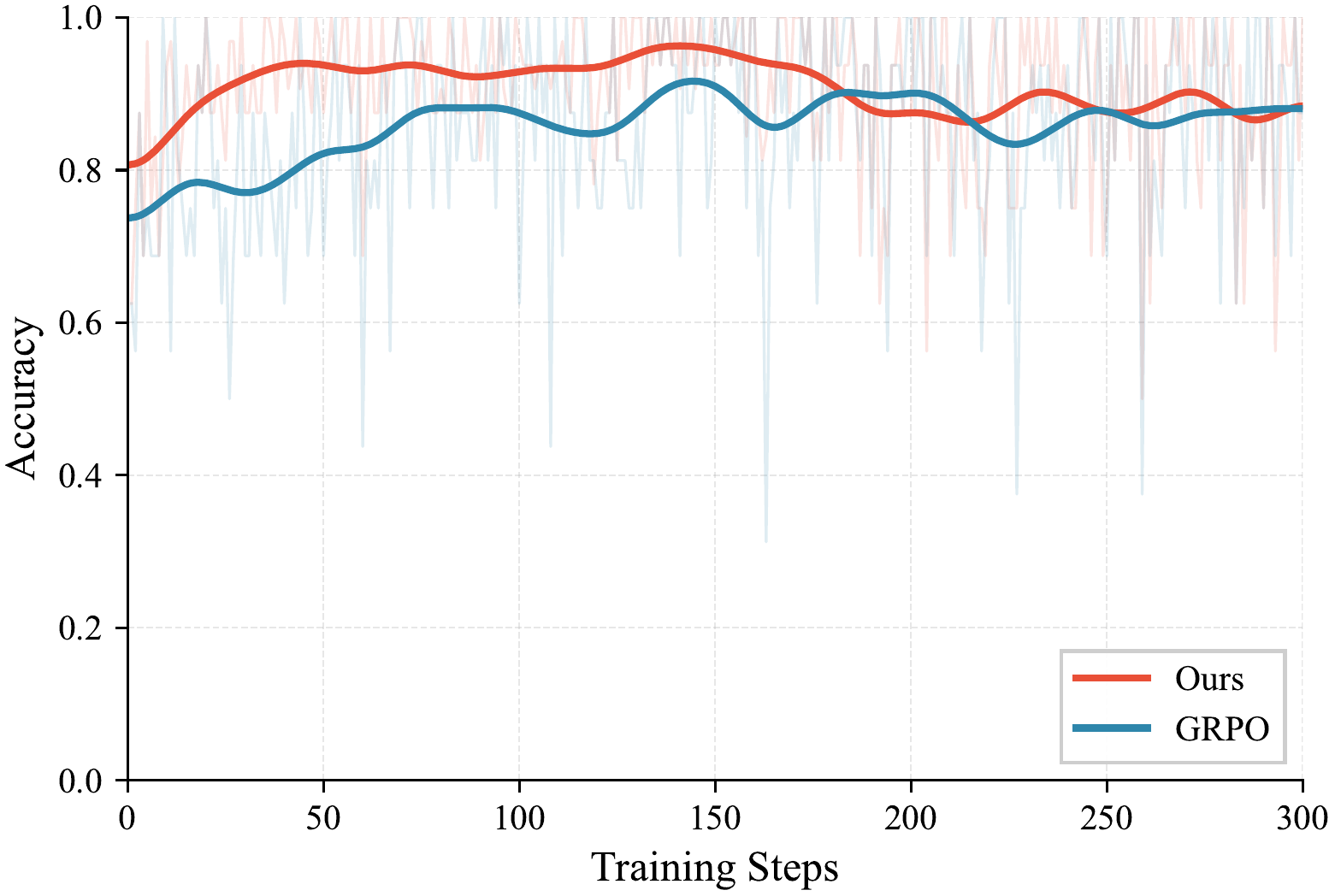}
    \small (a) GSM8K
\end{minipage}
\hfill
\begin{minipage}[t]{0.235\textwidth}
    \centering
    \includegraphics[width=\linewidth]{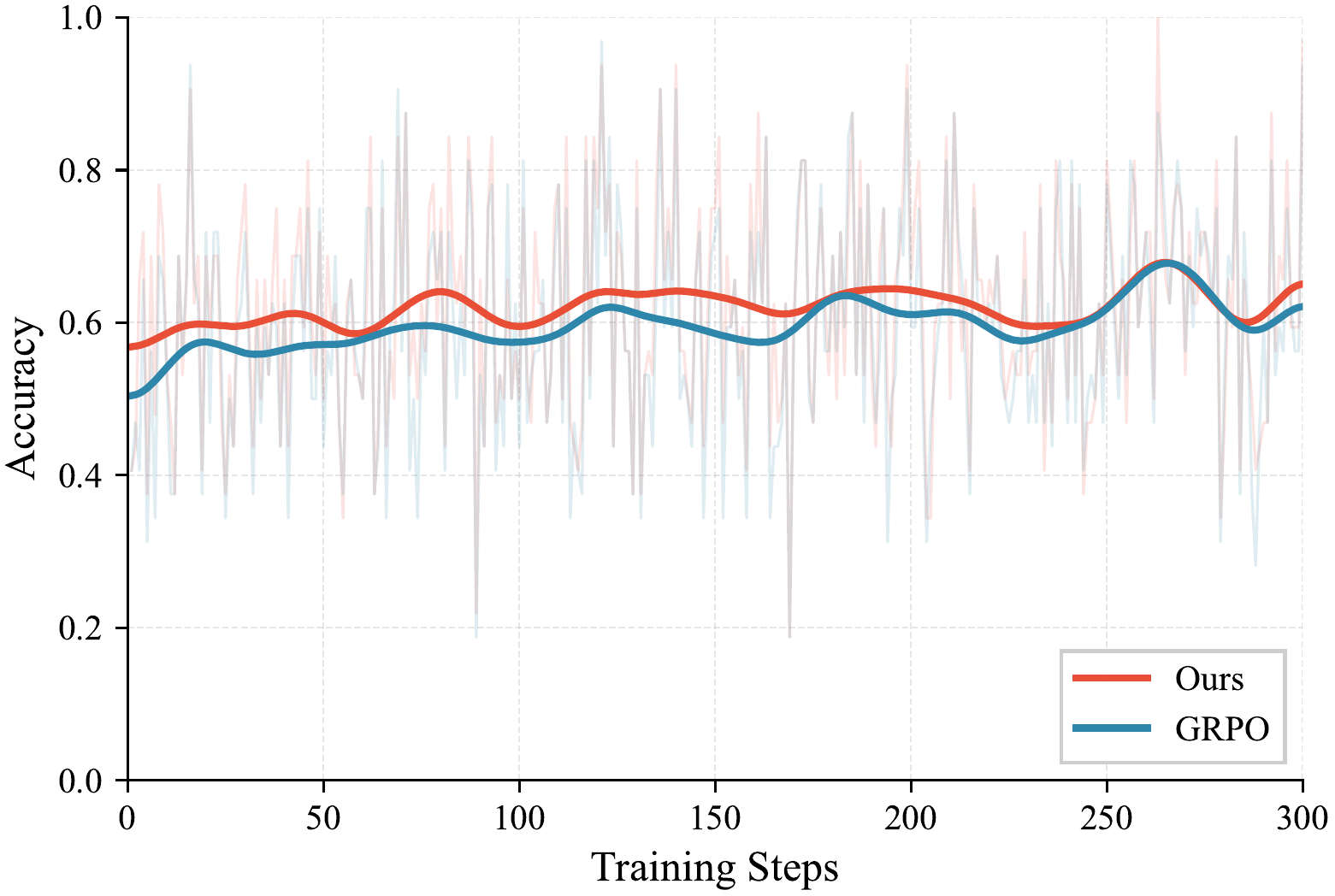}
    \small (b) MATH
\end{minipage}
\hfill
\begin{minipage}[t]{0.235\textwidth}
    \centering
    \includegraphics[width=\linewidth]{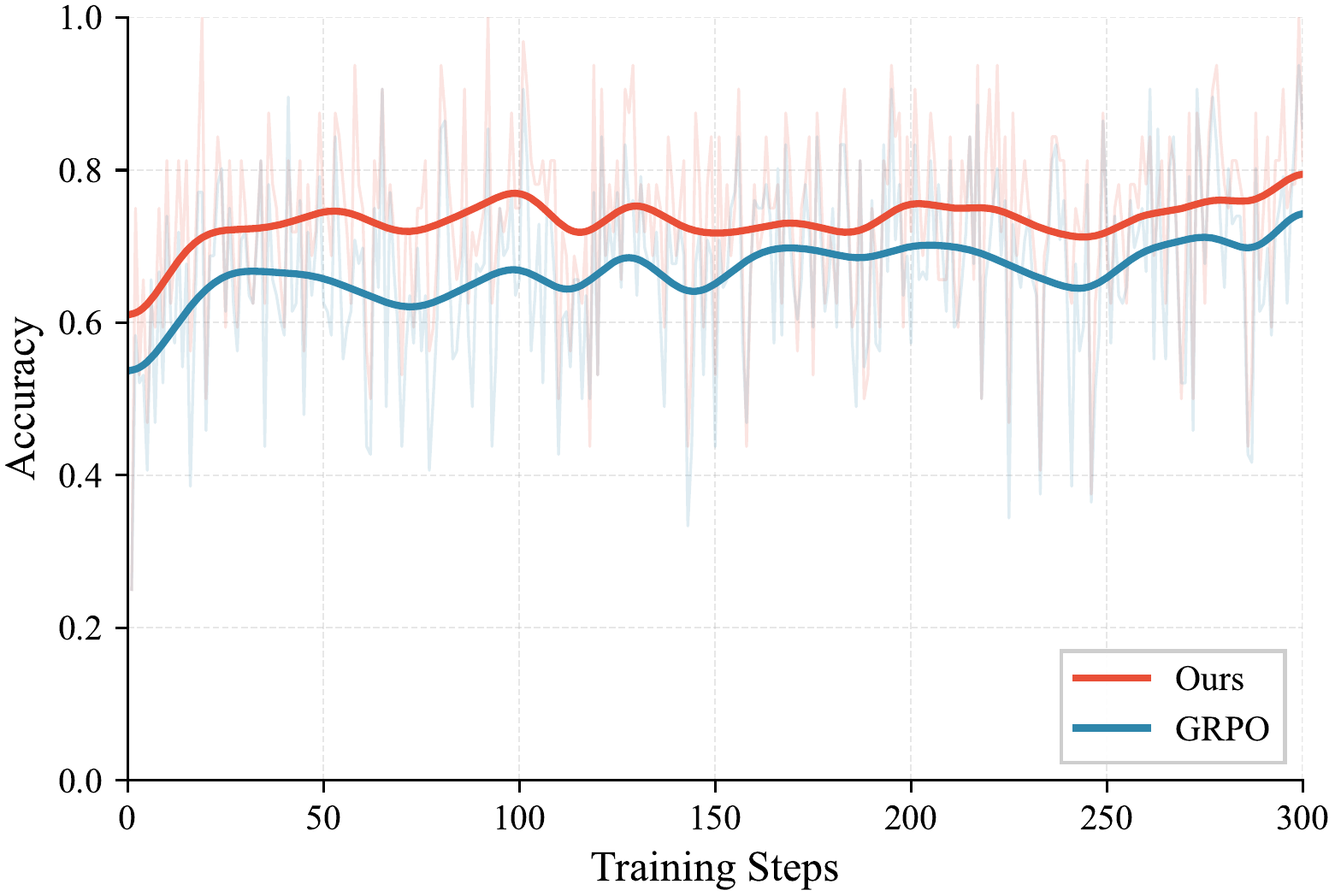}
    \small (c) Tal-SCQ5K-EN
\end{minipage}
\hfill
\begin{minipage}[t]{0.235\textwidth}
    \centering
    \includegraphics[width=\linewidth]{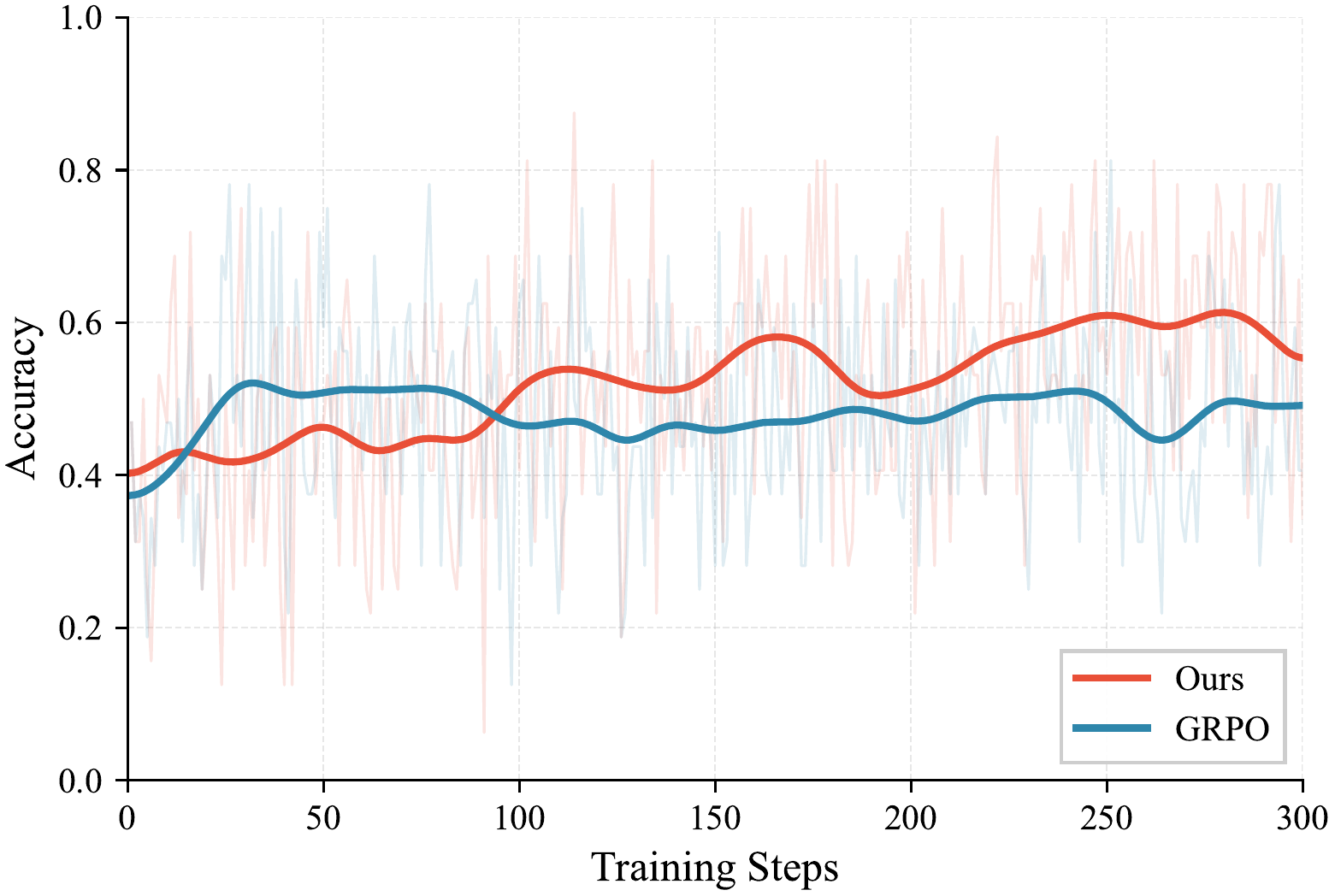}
    \small (d) Tal-SCQ5K-CN
\end{minipage}

\par\medskip

\begin{minipage}[t]{0.235\textwidth}
    \centering
    \includegraphics[width=\linewidth]{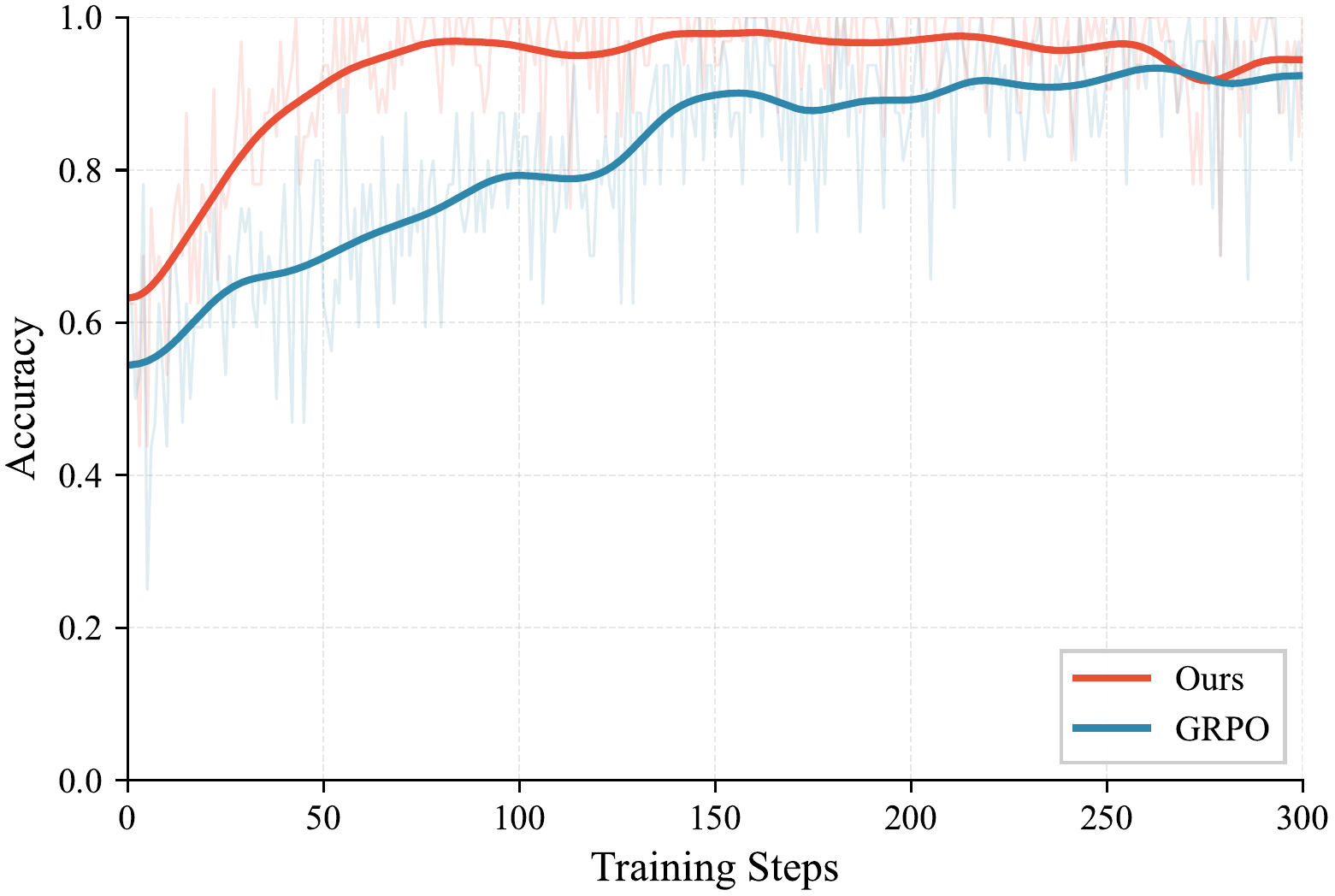}
    \small (e) Super-CLEVR
\end{minipage}
\hfill
\begin{minipage}[t]{0.235\textwidth}
    \centering
    \includegraphics[width=\linewidth]{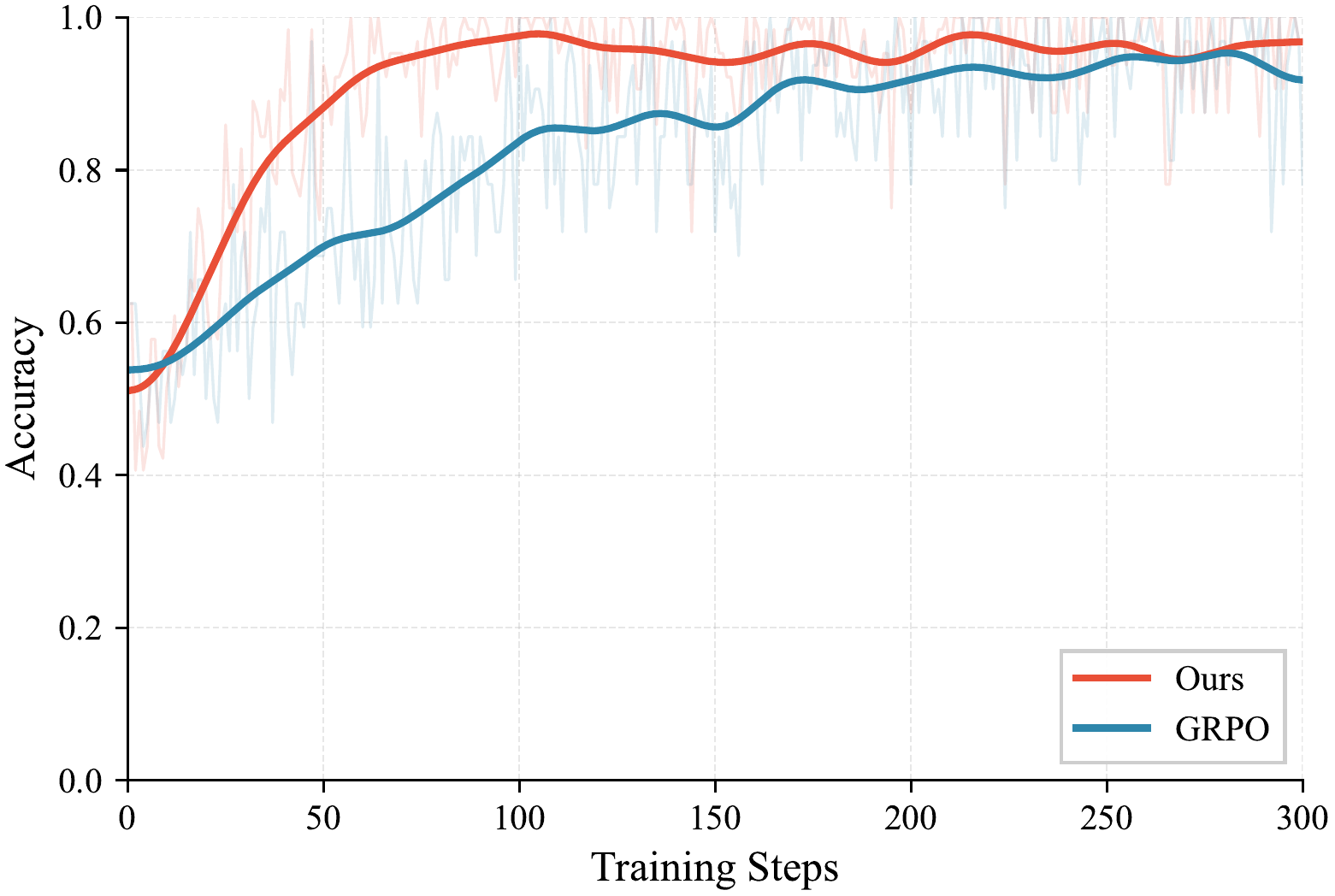}
    \small (f) CLEVR-CoGenT
\end{minipage}
\hfill
\begin{minipage}[t]{0.235\textwidth}
    \centering
    \includegraphics[width=\linewidth]{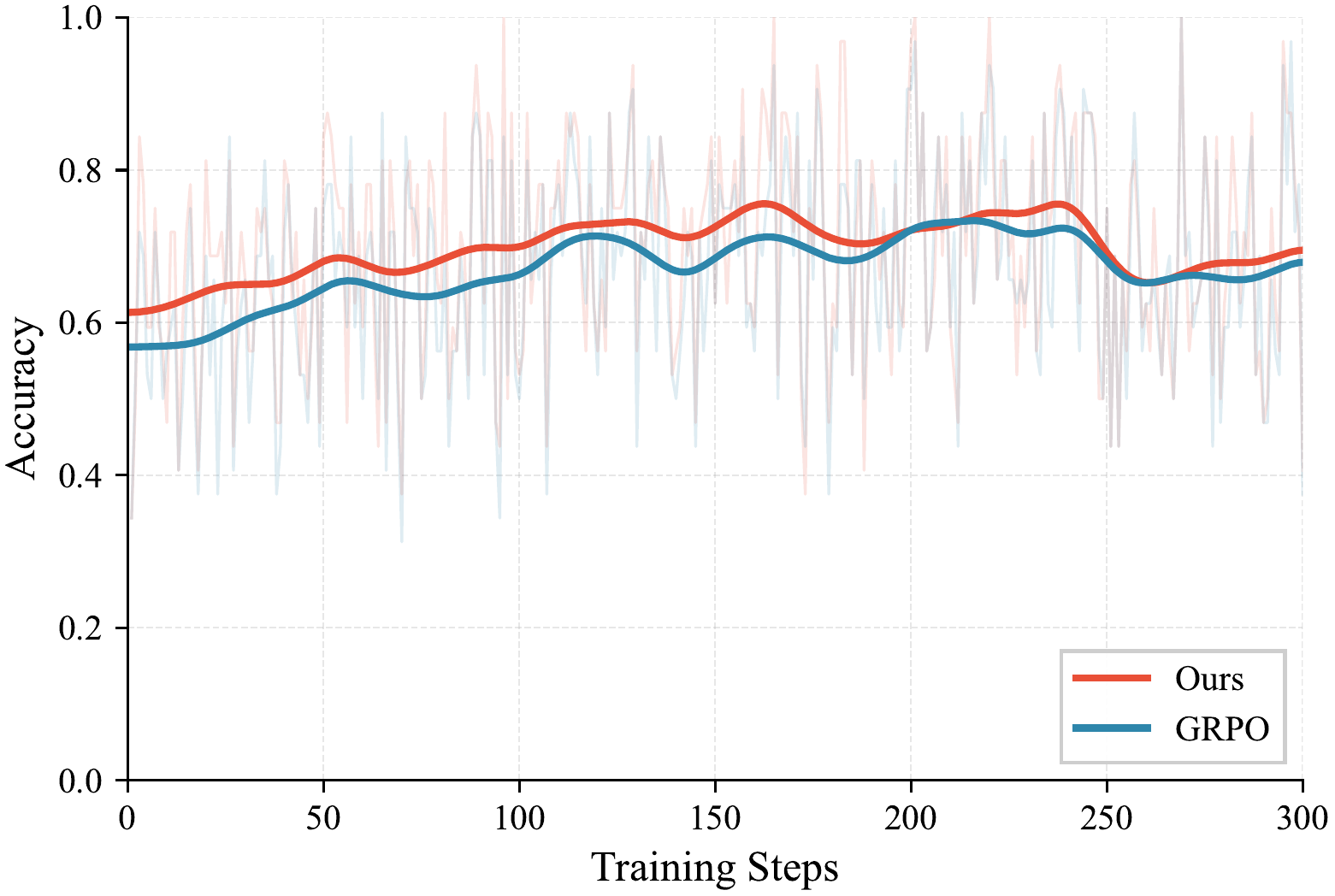}
    \small (g) ChartQA
\end{minipage}
\hfill
\begin{minipage}[t]{0.235\textwidth}
    \centering
    \includegraphics[width=\linewidth]{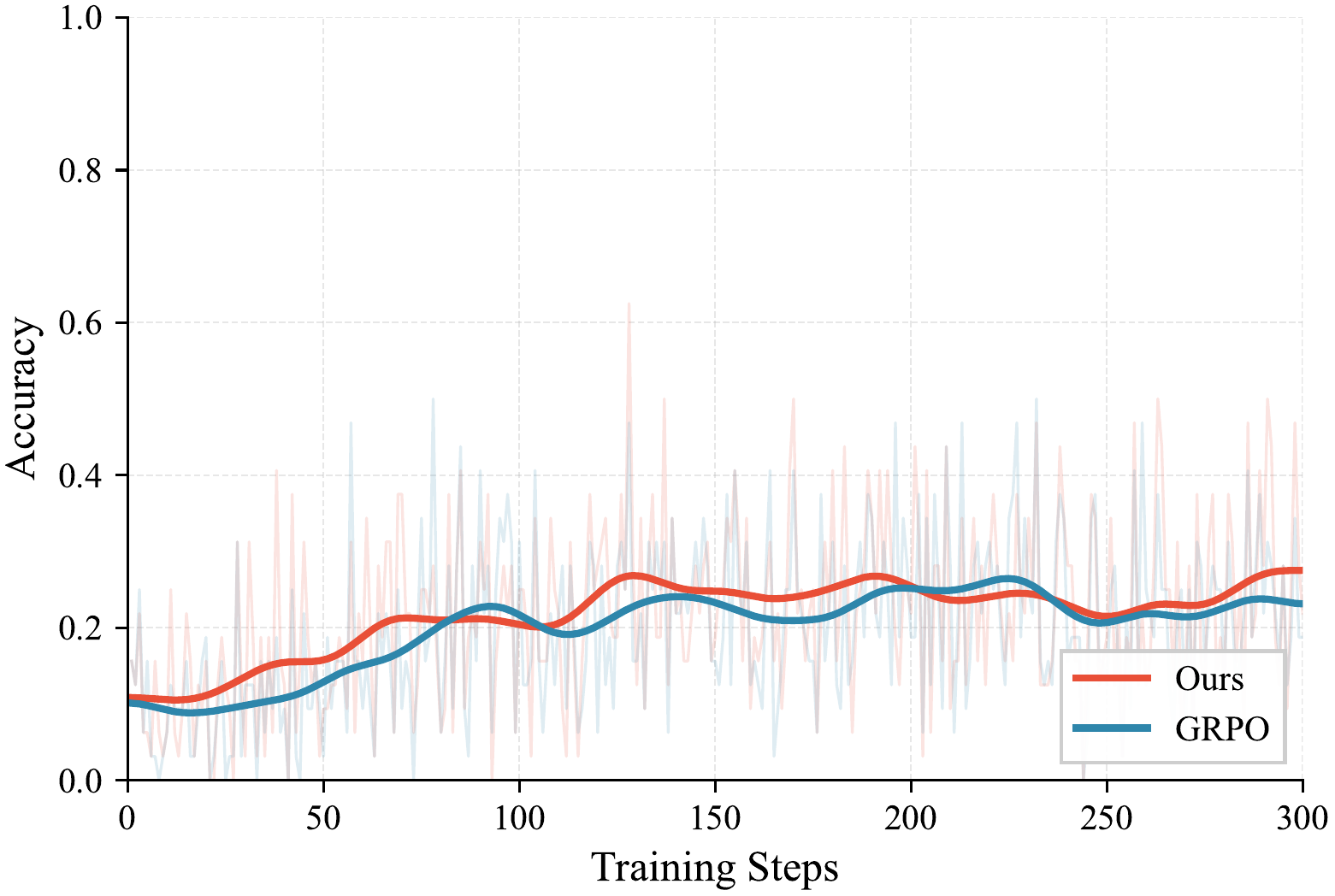}
    \small (h) CMM-Math
\end{minipage}
\caption{Training dynamics across text (top) and vision-language (bottom)
benchmarks. Faint lines are raw measurements and solid lines are
Gaussian-smoothed trends. Across most benchmarks, Ours reaches higher rollout
accuracy earlier than GRPO, with the clearest margins on GSM8K, Super-CLEVR,
and CLEVR-CoGenT.}
\label{fig:training_dynamics}
\end{figure*}

Across most benchmarks, Ours reaches higher rollout accuracy earlier in
training than GRPO, with the clearest margins on GSM8K, Super-CLEVR, and
CLEVR-CoGenT. The MATH, CMM-Math, and ChartQA curves overlap more strongly.
Figure~\ref{fig:training_dynamics} reports all eight task-specific runs. On
GSM8K, the advantage appears within the first 100 optimization steps. On
Super-CLEVR, Ours approaches saturation while GRPO remains in its initial
ascent. We do not claim a uniform separation on every task.
In the reported MATH and CMM-Math runs, GRPO also shows larger local
fluctuations, while the smoothed Ours trajectory is steadier over much of
training. This behavior is consistent with the HWM filtering sub-record
likelihood oscillations, although the curves alone do not establish that
mechanism.

\subsection{Broad-Data Transfer}

Table~\ref{tab:broad_transfer} tests whether the reward transfers beyond
task-specific training. On text, Ours improves over binary reward from 63.9 to
65.1 average accuracy at 3B and from 70.8 to 73.5 at 7B. At 3B, it attains the
highest average and the best GSM8K and CommonsenseQA scores; the SFT-free
variant is strongest on MATH-500. At 7B, Ours improves all four tasks over
binary reward. PRM-BoN@16 remains strongest overall, but uses 16 sampled
candidates and inference-time reranking.

For vision-language transfer, Ours improves the binary-reward average by 4.8
points at 3B and 1.8 points at 7B. At 3B, adding the visual gate improves the
average from 56.1 to 59.0, although the no-gate variant is 0.9 points higher on
ChartQA. At 7B, Ours attains the best average (65.5) and exceeds PRM-BoN@16 by
12.9 points. PRM-BoN@16 is strongest on MMStar, and binary reward is strongest
on ChartQA, while Ours leads on MathVista and MathVerse.

The scale comparison does not follow a single monotonic pattern. Relative to
binary reward, the text gain increases from 1.2 points at 3B to 2.7 at 7B,
whereas the vision-language gain decreases from 4.8 to 1.8 points. The method
therefore transfers across both tested scales, but these two points do not
establish a scaling law. They instead show that the effect remains observable
when model capacity and training hardware change under matched within-scale
comparisons.

\begin{table*}[t]
\centering
\small
\setlength{\tabcolsep}{3.0pt}
\begin{tabularx}{\textwidth}{@{}llYYYYY@{}}
\toprule
\multicolumn{7}{@{}l}{\textit{Text-only reasoning}} \\
\textbf{Scale} & \textbf{Method} &
\textbf{MMLU-Pro} & \textbf{GSM8K} & \textbf{MATH-500} &
\textbf{CSQA} & \textbf{Avg.} \\
\midrule
3B & Base & 35.8 & 78.8 & 56.0 & 69.6 & 60.1 \\
3B & Binary reward & \best{42.0} & 79.5 & 58.8 & 75.1 & 63.9 \\
3B & CoT-likelihood & 36.2 & 78.7 & 57.8 & 69.3 & 60.5 \\
3B & MIG only & 40.1 & 78.8 & 57.2 & 75.9 & 63.0 \\
3B & Redist-Prox & 41.0 & 80.2 & 60.6 & 75.6 & 64.4 \\
3B & Ours w/o SFT & 41.5 & 82.6 & \best{61.2} & 74.4 & 64.9 \\
\rowcolor{OursBlue}
3B & \ours{Ours} & \best{42.0} & \best{84.9} & 57.2 &
\best{76.2} & \best{65.1} \\
\midrule
7B & Base & 49.6 & 81.7 & 68.4 & 79.5 & 69.8 \\
7B & Binary reward & 52.1 & 82.1 & 69.4 & 79.4 & 70.8 \\
7B & PRM-BoN@16 & \best{58.8} & \best{95.5} & \best{76.2} &
79.4 & \best{77.5} \\
\rowcolor{OursBlue}
7B & \ours{Ours} & 54.7 & 87.7 & 70.4 & \best{81.3} & 73.5 \\
\midrule
\multicolumn{7}{@{}l}{\textit{Vision-language reasoning}} \\
\textbf{Scale} & \textbf{Method} &
\textbf{MathVista} & \textbf{MMStar} & \textbf{ChartQA} &
\textbf{MathVerse} & \textbf{Avg.} \\
\midrule
3B & Base & 55.4 & 46.2 & 79.9 & 20.3 & 50.5 \\
3B & Binary reward & 56.8 & 51.8 & 81.5 & 26.8 & 54.2 \\
3B & CoT-likelihood & 55.1 & 48.4 & 81.2 & 14.5 & 49.8 \\
3B & MIG only & 55.2 & 47.7 & 80.8 & 24.5 & 52.1 \\
3B & Ours w/o SFT & 55.3 & 48.5 & 81.6 & 26.6 & 53.0 \\
3B & Ours w/o visual gate & 59.1 & 50.0 & \best{81.8} & 33.6 & 56.1 \\
\rowcolor{OursBlue}
3B & \ours{Ours} & \best{62.1} & \best{53.5} & 80.9 &
\best{39.4} & \best{59.0} \\
\midrule
7B & Base & 52.0 & 42.3 & 86.3 & 26.8 & 51.9 \\
7B & Binary reward & 67.8 & 61.1 & \best{87.9} & 38.0 & 63.7 \\
7B & PRM-BoN@16 & 51.4 & \best{63.0} & 77.2 & 18.8 & 52.6 \\
\rowcolor{OursBlue}
7B & \ours{Ours} & \best{69.9} & 61.7 & 87.4 &
\best{42.8} & \best{65.5} \\
\bottomrule
\end{tabularx}
\caption{Broad-data transfer accuracy (\%). PRM-BoN@16 selects one answer
from 16 sampled completions; other methods use a single evaluated completion.}
\label{tab:broad_transfer}
\end{table*}

\subsection{What Drives the Transfer Gain?}

\paragraph{Step localization versus trace scoring.}
Conditional answer likelihood is not sufficient by itself. At 3B,
CoT-likelihood averages 60.5 on text and 49.8 on vision-language evaluation,
below the corresponding binary-reward averages of 63.9 and 54.2. Replacing
the single post-trace score with prefix-level HWM credit improves these
averages to 63.0 and 52.1 even in the MIG-only setting. The remaining gap to
the full method indicates that temporal localization and outcome grounding
play distinct roles: one identifies useful progress, while the other preserves
the requirement to finish correctly.

\paragraph{Outcome-gated self-distillation.}
On text transfer, removing gated SFT changes the average only from 65.1 to
64.9, but redistributes performance across tasks: the SFT-free variant is four
points higher on MATH-500, whereas the full method is 2.3 points higher on
GSM8K and 1.8 points higher on CommonsenseQA. On vision-language transfer, the
full objective improves more substantially over the SFT-free variant, from
53.0 to 59.0 average. Thus the auxiliary term is not uniformly beneficial for
every benchmark, but the result supports keeping it gated by both valid
structure and exact correctness rather than distilling every sampled trace.

\paragraph{Visual gate.}
The no-gate vision-language model reaches 56.1 average, compared with 59.0 for
the full method. The gate improves MathVista, MMStar, and MathVerse by 3.0,
3.5, and 5.8 points, respectively, while reducing ChartQA by 0.9 points. This
non-uniform result is important: the intervention is associated with stronger
overall transfer and with gains on three tasks, but it is not a universal
per-benchmark improvement. Controlled answer-changing examples in the
supplementary material provide complementary qualitative evidence about what
the real-versus-blank contrast captures.

\paragraph{Credit routing and inference cost.}
Redist-Prox is competitive on text transfer (64.4 average). It shares the
trajectory-level HWM mass and all objective weights with our SFT-free variant
and differs only in how step credit is routed, so the 0.5-point gap to Ours
w/o SFT (64.9) isolates the routing choice; the full method reaches 65.1.
Unlike PRM-BoN@16, both evaluate only one completion.
MIG therefore trades additional teacher-forced prefix scoring during training
for unchanged single-completion inference. The comparison to PRM-BoN@16 should
not be read as compute matched: BoN uses more test-time candidates, while MIG
uses more training-time scoring passes.

\subsection{Design Consequences}

\paragraph{Contrastive optimization and token assignment.}
Although $g_k$ and $R_{\mathrm{MIG}}$ are nonnegative, per-prompt group
normalization gives positive and negative $A^{\mathrm{step}}$ to trajectories
above and below the group mean; a constant process channel contributes no
update. Channel-wise normalization prevents raw reward scale alone from
suppressing the binary outcome channels.
Every prefix is probed, but only record-setting $g_k$ values enter the default
score $R_{\mathrm{MIG}}=\sum_k g_k$. Its group-relative advantage is then
applied over $M_{\mathrm{think}}$ rather than as an independently estimated
validity label for each step. Step decomposition determines the trajectory
preference, while masking keeps its process advantage off answer tokens.
Redist-Prox instead routes the same HWM mass to individual steps, providing the
matched routing ablation discussed above.

\paragraph{Why raw accumulation is retained.}
The raw sum preserves Proposition~\ref{prop:hwm_record}; dividing by $K$ would
change the score after inserting a below-watermark detour and invalidate
Corollary~\ref{cor:subrecord_detour}. This is not complete length invariance:
extra sub-record steps contribute nothing, but longer traces have more chances
to set a maximum through useful reasoning or score noise. Length bounds,
explicit step parsing, and terminal supervision constrain but do not eliminate
this exposure.

\paragraph{Scope of the potential interpretation.}
The potential-difference identity in Proposition~\ref{prop:hwm_record} is
algebraic: it explains why record improvements telescope on the
watermark-augmented trajectory. It does not imply policy invariance for the full
objective, which additionally applies group normalization, terminal rewards,
and gated self-distillation.

\paragraph{Grounding and distillation remain distinct.}
The SFT gates are computed from the unscaled format and correctness signals
before $S_{\mathrm{vis}}$ is applied to the RL channels. Thus, a structurally
valid, correct trajectory remains a distillation target even when the visual
gate attenuates its process and accuracy rewards. The real-versus-blank gap is
therefore a visual-sensitivity preference, not a relabeling of target
correctness.

\subsection{Scope of the Evidence}

MIG is on-policy and model-relative: prefix likelihoods are recomputed with the
current policy and detached before each update, so the same prefix can receive
a different score at a later checkpoint. A mathematically invalid step can
therefore receive credit if it raises reference-answer likelihood.
Group normalization removes prompt-group-wide additive shifts, but not
trajectory-specific miscalibration, motivating the separate terminal answer
channel. Likewise, a positive real-versus-blank gap indicates behavioral
sensitivity to the image intervention, but neither identifies the responsible
visual evidence nor establishes complete visual grounding. The visual-gate
ablation and controlled counterfactual examples test complementary
consequences of this signal. In the controlled examples, the question text is
fixed while answer-determining visual content is edited, so a correct answer
switch rules out a fixed text-only prediction for that instance. Because the
cases are selected and qualitative, they do not estimate population-level
counterfactual consistency. Task-specific experiments measure in-domain
optimization, whereas broad-data experiments measure transfer from WebData or
MMK12. Agreement across protocols, modalities, and scales broadens the evidence
but does not replace multi-seed uncertainty estimates. Given the single-run
design, the 0.5-point gap between Redist-Prox and Ours w/o SFT is an observed
ranking rather than a statistically resolved effect.

\section{Conclusion and Limitations}

MIG converts changes in reference-answer likelihood into dense process credit
without process labels or an auxiliary reward model. Across task-specific and
broad-data settings, its record-filtered process signal, outcome grounding, and
gated self-distillation improve over outcome-only training at 3B and 7B scales;
the visual gate improves average multimodal transfer, though not every task.
The method requires verifiable answers and explicit steps, and prefix scoring
adds training computation. Results are limited to single runs and models up to
7B. Multi-seed evaluation, more natural counterfactuals, flexible segmentation,
and adaptive prefix scoring remain open directions.

\clearpage
\appendix
\setcounter{figure}{0}
\setcounter{table}{0}
\setcounter{equation}{0}
\setcounter{algorithm}{0}
\setcounter{proposition}{0}
\renewcommand{\thefigure}{S\arabic{figure}}
\renewcommand{\thetable}{S\arabic{table}}
\renewcommand{\theequation}{S\arabic{equation}}
\renewcommand{\thealgorithm}{S\arabic{algorithm}}
\renewcommand{\theproposition}{S\arabic{proposition}}
\section*{Supplementary Material}
\section{Implementation Details}
\subsection{Step Parsing and Chunked MIG Computation}
\label{app:step_parsing_chunking}

\textbf{Format-driven step segmentation:}
Our stepwise reward computation relies on explicit step boundaries in the generated chain-of-thought.
To make these boundaries identifiable, we use a format-driven prompting strategy and require the model to generate reasoning in the following structure:
\begin{quote}
\small\ttfamily\raggedright\noindent
<think> \#\#\# Step 1 reasoning content $\cdots$ </think>
<answer>FINAL\_ANSWER</answer>
\end{quote}
During training, the parser searches the \texttt{<think>} span for step markers of the form \texttt{\#\#\# Step}.
Each segment between two adjacent step markers is treated as one reasoning step.
For the last step, the segment ends at the beginning of the \texttt{<answer>} span.

Formally, let $p_k$ denote the token position of the $k$-th step marker.
The token span of step $k$ is defined as
\[
s_k =
\begin{cases}
z_{p_k:p_{k+1}}, & k < K, \\
z_{p_k:p_{\mathrm{ans}}}, & k = K,
\end{cases}
\]
where $p_{\mathrm{ans}}$ is the token position at which the answer span begins.
If no valid step marker is found, the trajectory is assigned zero stepwise process reward and zero format reward.
This prevents degenerate solutions such as directly emitting an answer without structured reasoning.

\textbf{Robust marker matching:}
In practice, tokenization can make exact marker matching non-trivial, since byte-pair encoding may violate the simple decomposition
\[
\mathrm{encode}(A+B) \neq \mathrm{encode}(A) + \mathrm{encode}(B).
\]
To make step extraction robust to such boundary effects, we search for multiple marker variants, including \texttt{\#\#\# Step}, \texttt{\textbackslash n\#\#\# Step}, \texttt{ \#\#\# Step}, and variants with additional line breaks or spaces.
The parser operates at the token level and maps the matched marker positions back to completion token spans.
This implementation ensures that minor whitespace differences do not affect step boundary detection, while malformed trajectories without valid step markers still receive zero process credit.

\textbf{Per-step likelihood computation:}
For each parsed step $s_k$, we construct the reasoning prefix up to that step,
\[
c_k = (x, s_{1:k}),
\]
and compute the length-normalized log-likelihood of the ground-truth answer:
\[
\ell_k =
\frac{1}{|y^*|}
\sum_{t=1}^{|y^*|}
\log \pi_\theta
\left(
y^*_t \mid c_k, y^*_{<t}
\right).
\]
The MIG reward is then computed from the sequence
$\ell_0,\ell_1,\ldots,\ell_K$ using the monotonic historical watermark defined
in the main paper.

\textbf{Chunked reward computation:}
Naively computing MIG requires evaluating the answer likelihood for every trajectory-step pair.
For a rollout batch with $B$ prompts, $G$ generations per prompt, and an average of $\bar{K}$ parsed steps per generation, the number of step-conditioned likelihood evaluations is approximately
\[
B \times G \times \bar{K}.
\]
To control memory usage, we batch these trajectory-step pairs and evaluate them in small chunks.
Across model scales and hardware configurations, the chunk size ranges from 4
to 16; the 3B broad-data text configuration analyzed below uses a chunk size
of 8.
Chunking only affects throughput and peak memory usage; it does not change the reward definition or the resulting MIG values.
All prefixes are still evaluated exactly, and the final reward sequence is identical to what would be obtained by evaluating all trajectory-step pairs at once.

\subsection{Detailed Training Procedure}
\begin{algorithm}[t]
   \caption{Stepwise MIG Policy Optimization}
   \label{alg:mig_training_supp}
   \begin{algorithmic}[1]
      \STATE \textbf{Input:} dataset $\mathcal D$, policy $\pi_\theta$, group size $G$
      \REPEAT
      \STATE Sample prompts and $G$ completions per prompt
      \STATE Parse each completion into $s_{1:K}$ and answer $y$
      \FOR{each parsed completion $i$}
         \STATE Compute $\ell_0$; set $h_0\leftarrow\ell_0$ and $R_{\mathrm{MIG}}^{(i)}\leftarrow0$
         \FOR{$k=1,\ldots,K$}
            \STATE Compute $\ell_k$; set $g_k\leftarrow(\ell_k-h_{k-1})_+$
            \STATE $h_k\leftarrow\max(h_{k-1},\ell_k)$
            \STATE $R_{\mathrm{MIG}}^{(i)}\leftarrow R_{\mathrm{MIG}}^{(i)}+g_k$
         \ENDFOR
         \STATE Compute $r_{\mathrm{fmt}}^{(i)}$ and $r_{\mathrm{acc}}^{(i)}$; set the SFT gates
         \IF{the input contains an image}
            \STATE Compute $S_{\mathrm{vis}}^{(i)}$ from the real--blank likelihood gap
            \STATE Scale $R_{\mathrm{MIG}}^{(i)}$ and $r_{\mathrm{acc}}^{(i)}$ by $S_{\mathrm{vis}}^{(i)}$
         \ENDIF
      \ENDFOR
      \STATE Group-normalize each reward channel before combining them
      \STATE Compute the hybrid objective with decoupled reasoning and completion masks
      \STATE Update $\theta$ by gradient descent
      \UNTIL{training completes}
   \end{algorithmic}
\end{algorithm}
\subsection{Implementation and Hardware Protocol}
\label{app:implementation}

\textbf{Implementation:}
All experiments are implemented using PyTorch and DeepSpeed ZeRO-3 for memory optimization.

\textbf{Hardware Protocol:}
For 3B models, training is conducted on 4 NVIDIA A100 GPUs, and evaluation is conducted on a single A100 GPU.
For 7B models, training is conducted on 4 NVIDIA H200 GPUs, and evaluation is conducted on a single H200 GPU.
The task-specific in-domain experiments are conducted with 3B models, while the broad-data transfer experiments are conducted at both 3B and 7B scales.
For each dataset, model scale, and comparison group, all methods are trained and evaluated under the same hardware configuration to avoid hardware-induced variance.

\textbf{Training Configuration:}
Unless otherwise specified, we use the same training configuration across experiments.
The group size is set to $G=8$.
The maximum prompt length is set to 1024.
The maximum completion length is set to 2048 for 3B models and 4096 for 7B models.
During rollout, we use a constant sampling temperature of 1.0 to encourage exploration.
During evaluation, we use a sampling temperature of 0.6.

\textbf{Optimizer and Update Settings:}
We use AdamW with $(\beta_1,\beta_2)=(0.9,0.999)$,
$\epsilon=10^{-8}$, zero weight decay, a linear learning-rate schedule, and
global gradient-norm clipping at 10.0. The peak learning rate is
$10^{-6}$ for the task-specific and 7B experiments and $5\times10^{-7}$ for
the 3B broad-data experiments. For the latter, we use 30 warm-up update steps
for text-only training and 50 for vision-language training; the other runs use
no warm-up. Training uses BF16, a per-device prompt batch size of one, and
gradient accumulation of two for text-only runs, four for 3B vision-language
runs, and two for 7B vision-language runs. We set both the model and data seeds
to 42. The GRPO KL coefficient is 0 for all experiments.

\textbf{Objective Weights:}
The final training objective is weighted as
\[
\mathcal{L} = \lambda_{\mathrm{out}} \mathcal{L}_{\mathrm{Outcome}}
+ \lambda_{\mathrm{step}} \mathcal{L}_{\mathrm{MIG}}
+ \lambda_{\mathrm{sft}} \mathcal{L}_{\mathrm{Gated\text{-}SFT}} .
\]
Across all experiments, we set
$\lambda_{\mathrm{out}}=1.0$,
$\lambda_{\mathrm{step}}=0.01$, and
$\lambda_{\mathrm{sft}}=0.001$.
We select these coefficients by monitoring the gradient magnitudes induced by the three objective terms during training and balancing their effective update strengths.

\textbf{Visual Gate:}
For multimodal experiments, we use a global visual gate to down-weight reward signals for trajectories that do not depend strongly on the input image.
The default counterfactual image is a blank image.
We set the gate sharpness to $\tau=5.0$ and the margin to $m=0.1$.
For a fixed generated chain $z$, the gate is
\[
S_{\mathrm{vis}}=
\sigma\!\left(\tau[\ell_{\mathrm{real}}-\ell_{\mathrm{blank}}-m]\right),
\]
where $\ell_{\mathrm{real}}$ and $\ell_{\mathrm{blank}}$ are the length-normalized
reference-answer log-likelihoods under the real and blank images, respectively.
The gate is applied to the stepwise MIG reward and the binary accuracy reward, but not to the format reward or the gated SFT objective.

\subsection{Training-Time Compute Overhead}
\label{app:compute_overhead}

We measure the incremental cost of the text-only MIG likelihood probes in a
controlled 12-update Qwen2.5-3B-Instruct experiment on VeriFree WebData using
four NVIDIA A100 80GB GPUs. The control disables only the step-prefix and
prompt-only answer-likelihood probes; all shared training operations and
sampled completions remain matched. Table~\ref{tab:compute_overhead} reports
the mean over two timing runs per condition and a tensor-shape-based FLOPs
estimate for the additional probes.

\begin{table}[t]
\centering
\small
\caption{Controlled training-time overhead for text-only MIG at 3B scale.}
\label{tab:compute_overhead}
\setlength{\tabcolsep}{3.0pt}
\renewcommand{\arraystretch}{1.08}
\begin{tabularx}{\columnwidth}{@{}Xrrr@{}}
\toprule
\textbf{Metric} & \textbf{MIG off} & \textbf{MIG on} & \textbf{Change} \\
\midrule
Wall-clock (s/update) & 102.08 & 109.41 & $+7.17\%$ \\
Probe PFLOPs/update, no padding & 0 & 1.118 & $+1.118$ \\
Probe PFLOPs/update, chunk size 8 & 0 & 1.511 & $+1.511$ \\
\bottomrule
\end{tabularx}
\end{table}

As shown in Table~\ref{tab:compute_overhead}, MIG increases wall-clock time by
$7.17\%$. Its additional probes require approximately 1.118 padding-free
PFLOPs per update, or 1.511 PFLOPs with the within-chunk padding induced by our
chunk size of 8. FLOPs are estimated from the actual probe tensor shapes using
the standard dense-transformer approximation and exclude communication and
other systems overhead. The probes are used only during training, so MIG does
not increase single-completion inference cost.

\section{Proofs of Theoretical Properties}
\label{app:proofs}

\begin{proposition}[Record-Improvement Potential Difference]
\label{prop:hwm_record_supp}
Let $h_0=\ell_0$, $h_k=\max(h_{k-1},\ell_k)$, and
$g_k=(\ell_k-h_{k-1})_+$. Then
\[
h_k=\max_{0\leq j\leq k}\ell_j,\qquad
g_k=h_k-h_{k-1},
\]
and consequently
\[
\sum_{k=1}^{K}g_k
=\max_{0\leq j\leq K}\ell_j-\ell_0.
\]
Moreover, $g_k>0$ if and only if
$\ell_k>\max_{0\leq j<k}\ell_j$. With
$\tilde s_k=(x,s_{1:k},h_k)$ and $\Phi(\tilde s_k)=h_k$,
\[
g_k=\Phi(\tilde s_k)-\Phi(\tilde s_{k-1}).
\]
Thus, MIG is a potential difference on the watermark-augmented state.
\end{proposition}

\begin{corollary}[Invariance to Sub-Record Detours]
\label{cor:subrecord_detour_supp}
After any position $m<K$, insert a finite score sequence whose elements are all
at most the current watermark $h_m$, while leaving subsequent original scores
unchanged. Every inserted step receives zero reward and the total MIG reward is
unchanged.
\end{corollary}

\begin{proof}[Proof of Proposition~\ref{prop:hwm_record_supp}]
We prove $h_k=\max_{0\le j\le k}\ell_j$ by induction. The claim holds for
$k=0$ since $h_0=\ell_0$. Suppose it holds for $k-1$. Then
\[
    h_k
    =
    \max\{h_{k-1},\ell_k\}
    =
    \max\left\{\max_{0\le j<k}\ell_j,\ell_k\right\}
    =
    \max_{0\le j\le k}\ell_j .
\]
Thus the claim holds for all $k$.

Next, since
\[
    h_k=\max\{h_{k-1},\ell_k\},
\]
we have
\[
\begin{aligned}
    h_k-h_{k-1}
    &=
    \max\{h_{k-1},\ell_k\}-h_{k-1} \\
    &=
    \max\{0,\ell_k-h_{k-1}\}
    =
    (\ell_k-h_{k-1})_+
    =
    g_k .
\end{aligned}
\]
Summing over $k=1,\ldots,K$ gives the telescoping identity
\[
    \sum_{k=1}^{K}g_k
    =
    \sum_{k=1}^{K}(h_k-h_{k-1})
    =
    h_K-h_0 .
\]
Using $h_0=\ell_0$ and $h_K=\max_{0\le j\le K}\ell_j$, we obtain
\[
    R_{\mathrm{MIG}}
    =
    \max_{0\le j\le K}\ell_j-\ell_0 .
\]

The condition $g_k>0$ is equivalent to $\ell_k>h_{k-1}$. Since
\[
    h_{k-1}=\max_{0\le j<k}\ell_j,
\]
positive credit is assigned exactly to record-setting steps.

If $\ell_k\le 0$ for all $k$, then $h_K\le 0$. Since $h_K\ge h_0=\ell_0$,
we have $0\le h_K-h_0\le -\ell_0$, and therefore
\[
    0\le R_{\mathrm{MIG}}\le -\ell_0 .
\]
Finally, by the definition $\Phi(\tilde s_k)=h_k$, we have
\[
    \Phi(\tilde s_k)-\Phi(\tilde s_{k-1})
    =
    h_k-h_{k-1}
    =
    g_k .
\]
This proves the potential-difference representation.
\end{proof}

\begin{proof}[Proof of Corollary~\ref{cor:subrecord_detour_supp}]
Let $h_m$ be the watermark before the inserted detour. By assumption, every
inserted score satisfies $\tilde{\ell}_{m+r}\le h_m$. Therefore each inserted
step has gain
\[
    \tilde g_{m+r}
    =
    \max(0,\tilde\ell_{m+r}-h_m)
    =
    0,
\]
and the watermark remains $h_m$ throughout the detour.
After the detour, the subsequent original scores are unchanged and are compared
against the same watermark they would have faced without the detour. Hence all
future positive increments are unchanged, and the total reward
$R_{\mathrm{MIG}}$ is invariant to inserting any finite sub-record detour.
\end{proof}

\section{Prompt and Step Format}
\label{app:prompt_format}

\subsection{Instruction Template}

The policy is prompted to emit a parseable reasoning trace. The final answer must appear only inside an \texttt{<answer>} span, while the reasoning trace must appear inside a \texttt{<think>} span with explicit numbered step markers.
The following template is used in structured rollout:
\begin{quote}
\small\ttfamily\raggedright
\textbf{User:}\\
Camden just bought 3/4 times as many dogs as Rico, who has 10 more dogs than Justin. If Justin has 14 dogs, what's the total number of legs that Camden's dogs have?\\[2pt]
\textbf{System constraints:}\\
You MUST follow this exact output format (no extra text):\\
<think>\\
\#\#\# Step 1\\
...\\
\#\#\# Step 2\\
...\\
</think>\\
<answer>FINAL\_ANSWER\_HERE</answer>
\end{quote}

The strict formatting rules are: (1) the final answer must appear only inside \texttt{<answer>...</answer>}; (2) reasoning must be enclosed by \texttt{<think>...</think>}; (3) each reasoning unit must start with \texttt{\#\#\# Step X}; and (4) no extra text may appear outside the two tags.
Malformed trajectories without valid answer tags or step markers receive zero process reward and zero format reward.
In practice, we match marker variants with minor whitespace differences and map matched token positions back to completion spans, as described in the step-parsing subsection.

\subsection{Example Model Trajectory}

The following trajectory illustrates the stepwise output parsed by the structured rollout phase:
\begin{quote}
\small
\texttt{<think>}\\
\textbf{\#\#\# Step 1}\\
First, determine how many dogs Rico has. Rico has 10 more dogs than Justin, and Justin has 14 dogs. Thus Rico has $14+10$ dogs.\\
\textbf{\#\#\# Step 2}\\
Calculate the result: Rico has $24$ dogs.\\
\textbf{\#\#\# Step 3}\\
Camden bought $3/4$ times as many dogs as Rico, so Camden's dogs equal $3/4$ times Rico's dogs.\\
\textbf{\#\#\# Step 4}\\
Substitute Rico's number of dogs: Camden's dogs $=3/4\times 24$.\\
\textbf{\#\#\# Step 5}\\
Compute Camden's dogs: $(3\times 24)/4=18$.\\
\textbf{\#\#\# Step 6}\\
Each dog has 4 legs, so the total number of legs is the number of dogs times 4.\\
\textbf{\#\#\# Step 7}\\
Compute $18\times 4=72$.\\
\texttt{</think>}\\
\texttt{<answer>}72\texttt{</answer>}
\end{quote}

\section{Reward Proxy and Aggregation Details}
\label{app:proxy_details}

We empirically validate that the step-conditioned average log-likelihood $\ell_k$ serves as a useful proxy for the true value of an intermediate reasoning state.
We sample 100 problems from the GSM8K~\cite{cobbe_training_2021} test set, filter malformed trajectories and traces with fewer than two reasoning steps, and retain 84 problems with 225 intermediate states.
For each state, we estimate its rollout value with 32 Monte Carlo continuations at temperature 0.6, defining the value as the fraction of continuations that reach the correct final answer.

\begin{figure*}[t]
    \centering
    \begin{subfigure}{0.48\textwidth}
        \centering
        \includegraphics[width=\linewidth]{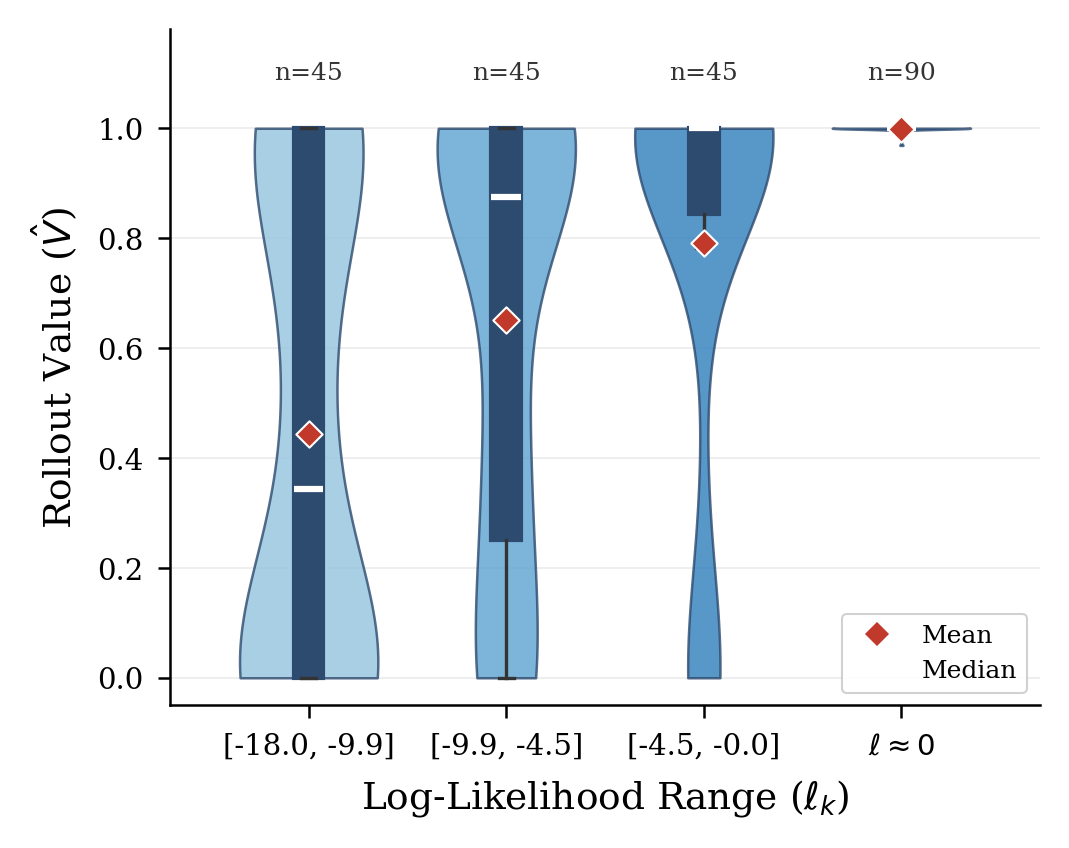}
        \caption{Value distribution vs. $\ell_k$}
        \label{fig:val_likelihood_app}
    \end{subfigure}
    \hfill
    \begin{subfigure}{0.48\textwidth}
        \centering
        \includegraphics[width=\linewidth]{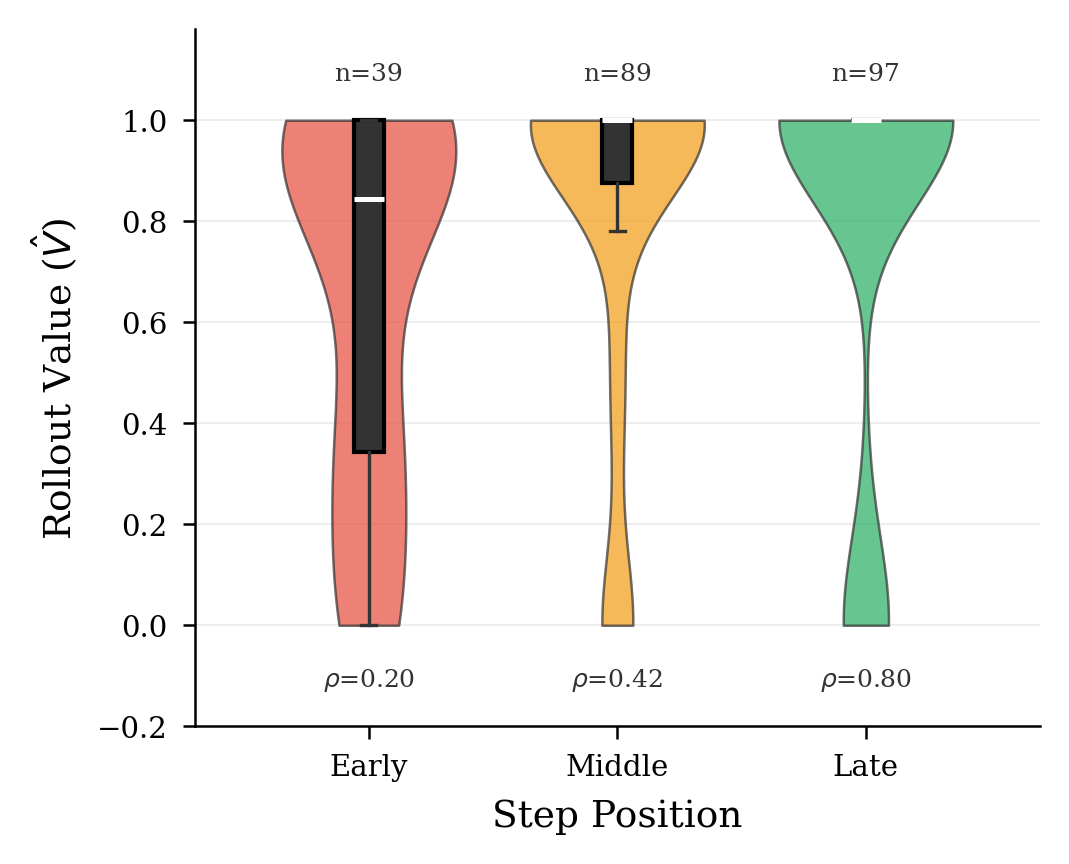}
        \caption{Value distribution vs. step position}
        \label{fig:val_position_app}
    \end{subfigure}
    \caption{Empirical validation of the reward proxy.
    Likelihood-conditioned values show a clearer monotonic trend than position-only grouping, supporting $\ell_k$ as a dense state-value proxy.}
    \label{fig:proxy_validation_plots}
\end{figure*}

\textbf{Visual analysis.}
Figure~\ref{fig:proxy_validation_plots} compares likelihood-conditioned value estimates with position-conditioned value estimates.
The likelihood buckets show a clear monotonic trend: as $\ell_k$ approaches zero, the rollout-estimated value increases sharply.
In contrast, position alone is a weaker heuristic. Late steps tend to have higher value, but early steps contain a wide mixture of poor and high-value states. This supports a content-aware signal rather than a fixed schedule that assumes later steps are always better.

\textbf{Baseline metrics.}
We compare $\ell_k$ with five alternatives: the raw likelihood increment $\ell_k-\ell_0$, relative position $k/T$, a binary last-step indicator, token count, and a random baseline.
Spearman correlation measures rank alignment with Monte Carlo rollout-estimated values, while ROC-AUC treats states with rollout value above 0.5 as positive examples.

\begin{table}[t]
    \centering
    \small
    \setlength{\tabcolsep}{2pt}
    \begin{tabular*}{\columnwidth}{@{\extracolsep{\fill}}lccc@{}}
        \toprule
        \textbf{Metric} & \textbf{Spearman $\rho$} & \textbf{ROC-AUC} & \textbf{p-value} \\
        \midrule
        \textbf{$\ell_k$ (Ours)} & \textbf{0.623} & 0.844 & $< 10^{-25}$ \\
        $\ell_k - \ell_0$ & 0.609 & \textbf{0.846} & $< 10^{-24}$ \\
        Relative Position ($k/T$) & 0.254 & 0.559 & $1.2 \times 10^{-4}$ \\
        Is Last Step & 0.176 & 0.526 & $8.3 \times 10^{-3}$ \\
        Token Count & 0.108 & 0.480 & $0.11$ \\
        Random & 0.026 & 0.507 & $0.70$ \\
        \bottomrule
    \end{tabular*}
    \caption{Reward proxy alignment with rollout-estimated values across 225
    intermediate states from GSM8K.}
    \label{tab:proxy_metrics}
\end{table}

\textbf{Proxy alignment.}
Table~\ref{tab:proxy_metrics} shows that $\ell_k$ achieves the strongest rank correlation among local state features ($\rho=0.623$), substantially outperforming relative position ($\rho=0.254$), token count ($\rho=0.108$), and the last-step heuristic ($\rho=0.176$).
The raw increment $\ell_k-\ell_0$ has similar ROC-AUC, but it lacks the monotonic filtering used by HWM and can still reward oscillatory trajectories.

\textbf{Reward aggregation strategies.}
Having established that $\ell_k$ is predictive of state value, we compare ways to aggregate the per-step likelihood sequence into a scalar reward.
The trajectory-level baselines are final likelihood $\ell_T$ and total delta $\ell_T-\ell_0$.
The dense step-level baselines are mean likelihood, sum likelihood, position-weighted mean (PWM), clipped delta, and our HWM reward
$\sum_k \max(0,\ell_k-h_{k-1})$.

\begin{table*}[t]
    \centering
    \small
    \setlength{\tabcolsep}{4pt}
    \begin{tabular*}{\textwidth}{@{\extracolsep{\fill}}llccc@{}}
        \toprule
        \textbf{Granularity} & \textbf{Aggregation} & \textbf{Spearman $\rho$} & \textbf{ROC-AUC} & \textbf{p-value} \\
        \midrule
        \multirow{2}{*}{Trajectory-level}
        & Final $\ell_T$ & 0.827 & 0.999 & $3.1 \times 10^{-22}$ \\
        & Delta Sum ($\ell_T-\ell_0$) & 0.670 & 0.982 & $3.1 \times 10^{-12}$ \\
        \midrule
        \multirow{5}{*}{Step-level}
        & \textbf{HWM (Ours)} & \textbf{0.671} & \textbf{0.982} & $2.9 \times 10^{-12}$ \\
        & PWM & 0.623 & 0.949 & $2.5 \times 10^{-10}$ \\
        & Mean $\ell$ & 0.600 & 0.932 & $1.6 \times 10^{-09}$ \\
        & Sum $\ell$ & 0.573 & 0.912 & $1.2 \times 10^{-08}$ \\
        & Clipped Delta & 0.411 & 0.799 & $1.0 \times 10^{-04}$ \\
        \bottomrule
    \end{tabular*}
    \caption{Comparison of reward aggregation strategies over the 84 retained
    trajectories. HWM is the strongest dense step-level aggregation method.}
    \label{tab:aggregation_comparison}
\end{table*}

\textbf{Aggregation analysis.}
Final $\ell_T$ has the highest correlation because it summarizes the completed trajectory, but it does not identify which intermediate step produced the improvement.
Among dense step-level alternatives, HWM is strongest ($\rho=0.671$, ROC-AUC $0.982$), matching the trajectory-level delta while distributing credit to record-setting steps.
PWM underperforms HWM, which empirically refutes the assumption that later reasoning steps should be rewarded simply because of their position.
Mean and sum likelihood are more sensitive to low-confidence oscillations, whereas HWM filters them through the historical maximum.

\section{Additional Ablation Results}
\label{app:additional_ablations}

\subsection{Effectiveness of Answer Variants}
\label{app:answer_variants}

Mathematical answers may have multiple equivalent surface forms, such as ``1.6'' and ``8/5''.
When computing step-conditioned likelihood, using only the canonical answer string can therefore introduce false negative reward signals for trajectories that support an equivalent answer form.
To reduce this mismatch, the variant-aware setting uses an augmented answer set $\mathcal{Y}^*$ and takes the maximum likelihood over candidate variants:
\begin{equation}
    \ell_k = \max_{y \in \mathcal{Y}^*}
    \left(
    \frac{1}{|y|}
    \sum_{t=1}^{|y|}
    \log \pi_\theta(y_t \mid x, s_{1:k}, y_{<t})
    \right).
\end{equation}
We use a processed 7,500-example MATH~\cite{hendrycks_measuring_2021} training split containing up to five semantically equivalent short-answer forms per problem.
We generate the answer variants offline using Qwen2.5-32B-Instruct~\cite{qwen25_2024}.
The processed dataset used in our experiments is provided in the supplementary material.

\begin{table}[t]
\centering
\small
\begin{tabular*}{0.82\columnwidth}{@{\extracolsep{\fill}}lc@{}}
\toprule
\textbf{Setting} & \textbf{MATH Pass@1} \\
\midrule
Canonical answer only & 60.5 \\
Variant-aware likelihood & \textbf{61.4} \\
\bottomrule
\end{tabular*}
\caption{Effect of answer variants on MATH Pass@1 accuracy (\%).}
\label{tab:answer_variants}
\end{table}

As shown in Table~\ref{tab:answer_variants}, variant-aware likelihood improves MATH Pass@1 from $60.5\%$ to $61.4\%$.
The gain is modest but consistent with the hypothesis that equivalence-aware scoring reduces surface-form mismatch.

\subsection{Difficulty-Level Breakdown on MATH}
\label{app:math_breakdown}

Table~\ref{tab:math_difficulty_breakdown} breaks down MATH performance by difficulty level.
Both Ours configurations use the variant-aware answer likelihood described in the answer-variants subsection; the full model therefore has an overall MATH Pass@1 accuracy of $61.4\%$, rather than the canonical-answer result of $60.5\%$ reported in the task-specific main table.
On Level 1, the full method reaches $86.0\%$, compared with $79.1\%$ for Ours without SFT, $74.4\%$ for GRPO, and $81.4\%$ for the base model.
On Level 5, the SFT-free variant is slightly stronger ($35.1\%$ vs. $33.6\%$), while the full hybrid objective gives the highest overall accuracy.

\begin{table}[t]
\centering
\small
\setlength{\tabcolsep}{2pt}
\begin{tabular*}{\columnwidth}{@{\extracolsep{\fill}}lrrrrrr@{}}
\toprule
\textbf{Method} & \textbf{L1} & \textbf{L2} & \textbf{L3} & \textbf{L4} & \textbf{L5} & \textbf{Overall} \\
\midrule
Base Model & 81.4 & 76.7 & 64.8 & 57.0 & 26.1 & 56.0 \\
GRPO & 74.4 & 75.6 & 71.4 & 54.7 & 33.6 & 58.0 \\
$\text{Ours}_{\text{w/o SFT}}$ & 79.1 & \textbf{78.9} & \textbf{77.1} & 55.5 & \textbf{35.1} & 60.8 \\
\textbf{Ours} & \textbf{86.0} & 77.8 & \textbf{77.1} & \textbf{57.8} & 33.6 & \textbf{61.4} \\
\bottomrule
\end{tabular*}
\caption{Performance by difficulty level on Hendrycks MATH. Both Ours rows use
variant-aware likelihood. Overall accuracy is weighted by the number of
examples at each level.}
\label{tab:math_difficulty_breakdown}
\end{table}

\subsection{Problem-Type Breakdown on SVAMP}
\label{app:svamp_breakdown}

To examine out-of-distribution arithmetic behavior, we evaluate the same 3B model variants used in the MATH analysis on the full SVAMP~\cite{patel2021nlp} test split ($n=300$) without additional SVAMP training and report exact-answer Pass@1 accuracy.
Both Ours configurations use variant-aware answer likelihood.
We categorize the SVAMP examples by problem type in Table~\ref{tab:svamp_type_breakdown}.
The results mirror the MATH difficulty trade-off.
Compared with the SFT-free variant, the full method is more accurate on Subtraction, Multiplication, and Addition, but less accurate on Common-Divisor problems.

\begin{table}[t]
\centering
\small
\setlength{\tabcolsep}{2pt}
\begin{tabular*}{\columnwidth}{@{\extracolsep{\fill}}lrrrr@{}}
\toprule
\textbf{Problem type} & \textbf{Base} & \textbf{GRPO} &
\shortstack{\textbf{Ours}\\\textbf{w/o SFT}} & \textbf{Ours} \\
\midrule
Subtraction & 88.8 & 88.8 & 87.5 & \textbf{90.6} \\
Common-Divisor & 79.2 & 75.0 & \textbf{83.3} & 77.1 \\
Multiplication & 75.8 & 78.8 & 78.8 & \textbf{90.9} \\
Addition & 84.7 & 86.4 & 89.8 & \textbf{91.5} \\
\bottomrule
\end{tabular*}
\caption{Exact-answer Pass@1 accuracy (\%) by SVAMP problem type. Both Ours
rows use variant-aware answer likelihood.}
\label{tab:svamp_type_breakdown}
\end{table}

\subsection{Validation of Monotonic Accumulation}
\label{app:monotonic_accumulation}

\begin{figure}[t]
    \centering
    \includegraphics[width=\columnwidth]{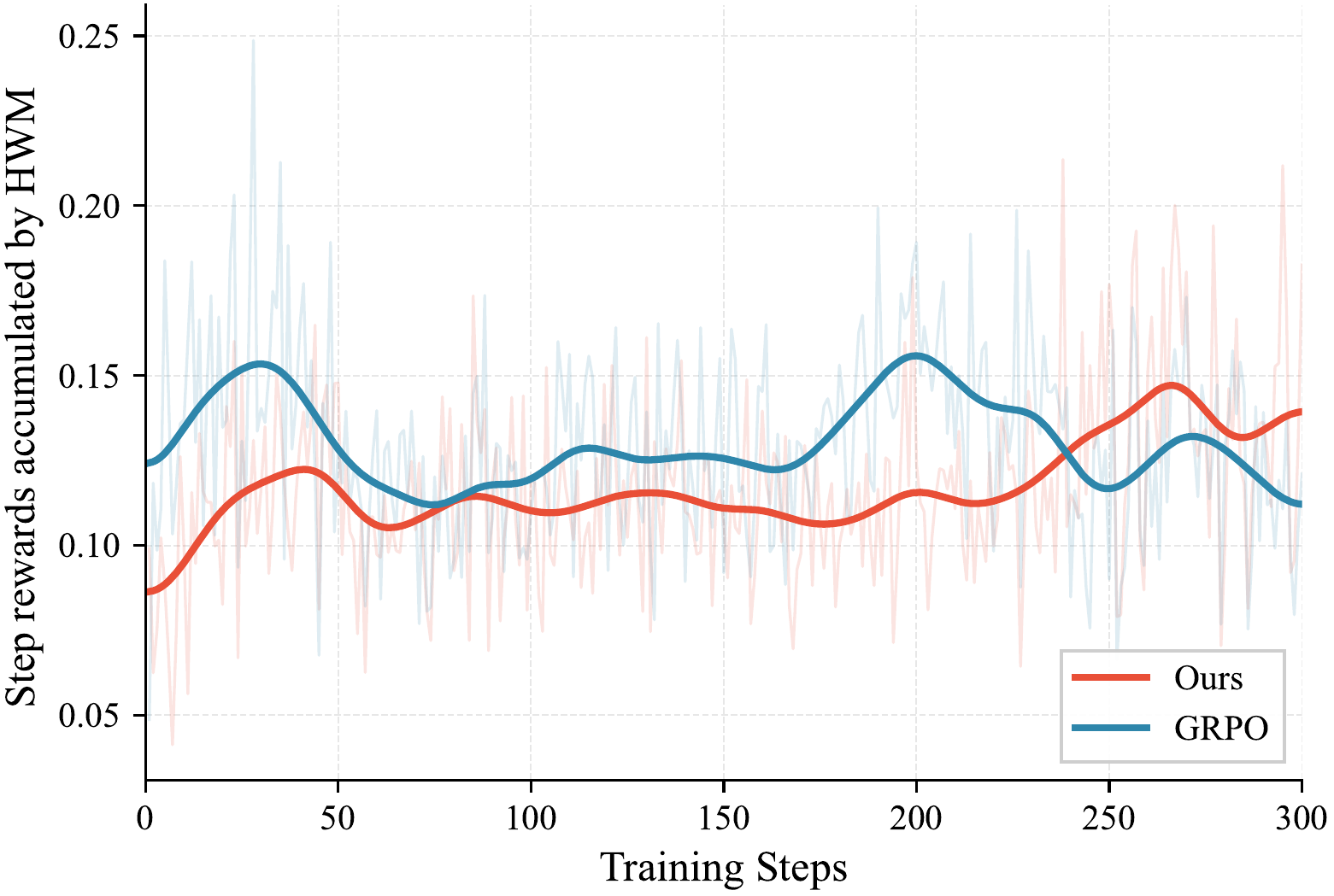}
    \caption{Accumulated HWM reward during training. Ours exhibits a positive overall trend and finishes above GRPO, whereas GRPO shows no sustained upward trend.}
    \label{fig:step_reward_evolution}
\end{figure}

Figure~\ref{fig:step_reward_evolution} compares the accumulated HWM reward throughout training.
Ours exhibits a positive overall trend, whereas GRPO fluctuates without a sustained upward trend.
By the end of training, Ours attains a higher accumulated HWM reward, consistent with MIG increasingly favoring trajectories that contain record-setting likelihood improvements.

\section{Qualitative Examples}
\label{app:qualitative_examples}

We provide two illustrative MATH examples and one multimodal case study containing controlled ChartQA and Super-CLEVR counterfactuals to examine the behavior of dense stepwise rewards and visual gating.
Unlike the summarized discussion in the main text, this appendix keeps the original problem statements, step-level summaries, and visual intervention used in the qualitative analysis.
The first case illustrates a trajectory in which Ours avoids premature stopping on a recursive reasoning problem.
The second illustrates how a more granular trajectory can increase exposure to arithmetic errors on a simpler algebraic problem.
The third examines whether model predictions track answer-changing edits in chart reading and object-centric reasoning.

\subsection{Case Study 1: Deep Reasoning Capability (Level 5)}
\label{app:case1_success}

\textbf{Original problem.}
\begin{quote}
\small
The proper divisors of 12 are 1, 2, 3, 4 and 6.
A proper divisor of an integer $N$ is a positive divisor of $N$ that is less than $N$.
What is the sum of the proper divisors of the sum of the proper divisors of 284?
\end{quote}
\textbf{Correct answer:} 284. The intermediate values 220 and 284 are an amicable pair.

\begin{table}[t]
\centering
\small
\begin{tabular}{@{}>{\raggedright\arraybackslash}p{0.26\columnwidth}>{\raggedright\arraybackslash}p{0.68\columnwidth}@{}}
\toprule
\multicolumn{2}{@{}l@{}}{\textbf{\textcolor{teal}{MIG (Ours) -- Correct}}} \\
\midrule
\textbf{Trace length} & \textit{Total 20 steps, 941 tokens.} \\
\textbf{Step 1--8} & Identifies proper divisors of 284: $\{1,2,4,71,142\}$. \\
\textbf{Step 9--11} & Calculates the first sum: $1+2+4+71+142=220$. \\
\textbf{Step 12--15} & \textcolor{blue}{Crucial transition.} The model recognizes 220 as an intermediate result and finds the proper divisors of 220: $\{1,2,4,5,10,11,20,22,44,55,110\}$. \\
\textbf{Step 16--19} & Calculates the second sum: $1+2+\cdots+110=284$. \\
\textbf{Step 20} & Final answer: 284. \\
\textbf{Analysis} & The Ours trajectory completes both divisor-sum operations rather than terminating at the intermediate value 220. Its continuation beyond the first milestone is consistent with MIG favoring prefixes that produce further record-setting gains in answer likelihood. \\
\midrule
\multicolumn{2}{@{}l@{}}{\textbf{\textcolor{red}{GRPO (Baseline) -- Incorrect}}} \\
\midrule
\textbf{Trace length} & \textit{Total 7 steps, 666 tokens.} \\
\textbf{Step 1} & Identifies proper divisors of 284 correctly. \\
\textbf{Step 2} & Calculates the first sum as 220. \\
\textbf{Step 3} & Identifies proper divisors of 220, but does not reliably continue the nested objective. \\
\textbf{Step 4--5} & \textcolor{red}{Computation collapse.} Instead of summing the divisors of 220, the model loses track of the second layer. \\
\textbf{Step 6--7} & Prematurely concludes with the intermediate value or a hallucinated sum. Final answer: 220. \\
\textbf{Analysis} & The GRPO trajectory reaches the intermediate value 220 but fails to complete the second divisor-sum operation, illustrating a premature-stopping failure under outcome-only training. \\
\bottomrule
\end{tabular}
\caption{Case Study 1: comparison on multi-hop divisor reasoning. The problem
requires two nested divisor-sum computations. Ours completes both, whereas
GRPO stops at the intermediate result.}
\label{tab:qualitative_case_divisor}
\end{table}

\subsection{Case Study 2: The Cost of Granularity (Level 3)}
\label{app:case2_failure}

\textbf{Original problem.}
\begin{quote}
\small
The area of triangle $ABC$ is equal to $a^2-(b-c)^2$.
Compute $\tan A$.
\end{quote}
\textbf{Correct answer:} $8/15$.

\begin{table}[t]
\centering
\small
\begin{tabular}{@{}>{\raggedright\arraybackslash}p{0.26\columnwidth}>{\raggedright\arraybackslash}p{0.68\columnwidth}@{}}
\toprule
\multicolumn{2}{@{}l@{}}{\textbf{\textcolor{red}{MIG (Ours) -- Incorrect}}} \\
\midrule
\textbf{Trace length} & \textit{Total 19 steps, 1199 tokens.} \\
\textbf{Step 1--5} & Expands $a^2-(b-c)^2$ and equates it to $\frac{1}{2}bc\sin A$. The model correctly derives $bc\sin A=4bc(1-\cos A)$. \\
\textbf{Step 6--16} & \textcolor{red}{Over-decomposition.} The model breaks the quadratic equation solving process into many small arithmetic substeps. \\
\textbf{Step 17} & \textit{Calculation error.} The trace reaches an expression of the form ``solving $17x^2-32x+15=0$'', but a sign error occurs in the explicit quadratic-formula expansion. \\
\textbf{Step 18--19} & The arithmetic error is propagated to the final result. Final answer: 4. \\
\textbf{Analysis} & The Ours trajectory decomposes the solution into more steps and then makes a local arithmetic error. This example illustrates that fine-grained reasoning can increase exposure to local mistakes even when the overall derivation is well motivated. \\
\midrule
\multicolumn{2}{@{}l@{}}{\textbf{\textcolor{teal}{GRPO (Baseline) -- Correct}}} \\
\midrule
\textbf{Trace length} & \textit{Total 5 steps, 1060 tokens.} \\
\textbf{Step 1--2} & Uses the cosine rule more directly and arrives at $\sin A=4(1-\cos A)$. \\
\textbf{Step 3--4} & Efficiently solves the quadratic equation for $\cos A$. \\
\textbf{Step 5} & Calculates $\tan A=\frac{8/17}{15/17}=8/15$. \\
\textbf{Answer} & $8/15$. \\
\textbf{Analysis} & For this simpler procedural problem, the concise path reduces arithmetic exposure and reaches the correct answer. \\
\bottomrule
\end{tabular}
\caption{Case Study 2: comparison on algebraic manipulation. Ours follows a
longer derivation and makes an arithmetic error; GRPO reaches the answer with
a shorter derivation.}
\label{tab:qualitative_case_triangle}
\end{table}

\subsection{Case Study 3: Counterfactual Visual Grounding}
\label{app:case3_visual_grounding}

\begin{figure*}[t]
    \centering
    {\small\textbf{ChartQA question:} What age group has the highest percentage of Associate degrees?\par}
    \vspace{0.3em}
    \begin{subfigure}[t]{0.48\textwidth}
        \centering
        \includegraphics[width=\linewidth]{Figs/chartqa_visual_gate_original.png}
        \caption{Original chart (GT: 25--34).}
        \label{fig:visual_gate_chartqa_original}
    \end{subfigure}
    \hfill
    \begin{subfigure}[t]{0.48\textwidth}
        \centering
        \includegraphics[width=\linewidth]{Figs/chartqa_visual_gate_counterfactual.png}
        \caption{Counterfactual: red-bar maximum changed (GT: 35--44).}
        \label{fig:visual_gate_chartqa_counterfactual}
    \end{subfigure}

    \vspace{0.7em}
    {\small\textbf{Super-CLEVR question:} How many different items are there in the image?\par}
    \vspace{0.3em}
    \begin{subfigure}[t]{0.48\textwidth}
        \centering
        \includegraphics[width=\linewidth]{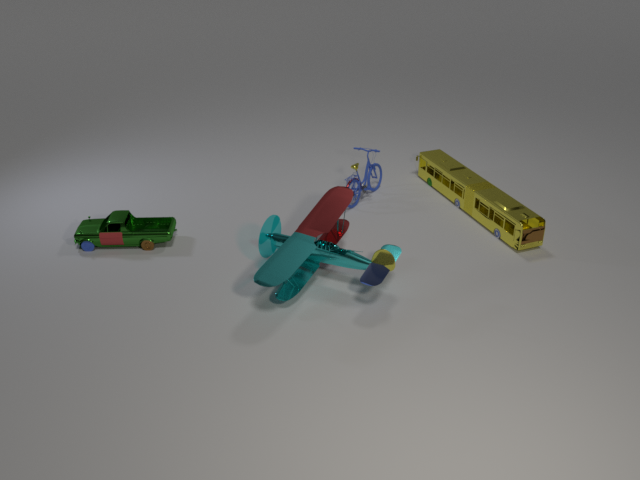}
        \caption{Original scene (GT: 4).}
        \label{fig:visual_gate_superclevr_original}
    \end{subfigure}
    \hfill
    \begin{subfigure}[t]{0.48\textwidth}
        \centering
        \includegraphics[width=\linewidth]{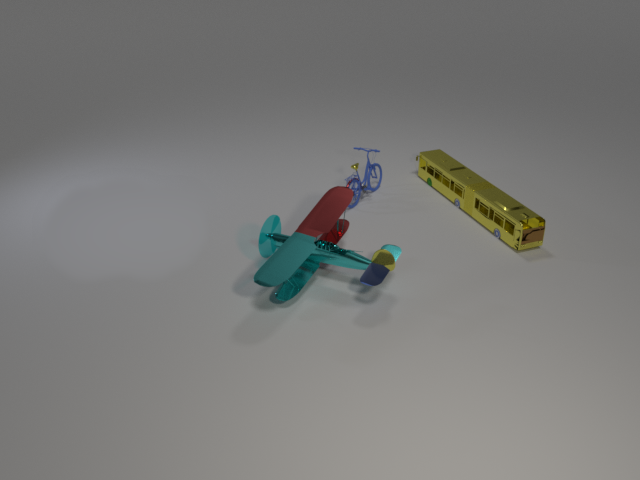}
        \caption{Counterfactual: green truck removed (GT: 3).}
        \label{fig:visual_gate_superclevr_counterfactual}
    \end{subfigure}

    \vspace{0.5em}
    \small
    \begin{tabular}{@{}llccc@{}}
        \toprule
        \textbf{Benchmark (GT change)} & \textbf{Method} & \textbf{Original} & \textbf{Counterfactual} & \textbf{Blank} \\
        \midrule
        \multirow{2}{*}{ChartQA (25--34 $\rightarrow$ 35--44)}
            & Ours w/o visual gate & 25--34 $\checkmark$ & 25--34 $\times$ & 25--34 \\
            & Ours & 25--34 $\checkmark$ & 35--44 $\checkmark$ & 25--34 \\
        \midrule
        \multirow{2}{*}{Super-CLEVR ($4 \rightarrow 3$)}
            & Ours w/o visual gate & $4\;\checkmark$ & $4\;\times$ & 3 \\
            & Ours & $4\;\checkmark$ & $3\;\checkmark$ & 3 \\
        \bottomrule
    \end{tabular}
    \caption{Case Study 3: controlled counterfactual visual grounding.
    Each row changes only the visual evidence that determines the answer while keeping the question fixed. Both methods answer the original inputs correctly. Ours follows both counterfactual changes, whereas the variant without the visual gate retains its original predictions. Blank-image outputs are diagnostic and have no ground-truth label.}
    \label{fig:qualitative_visual_gate}
\end{figure*}

We construct two controlled counterfactual pairs from ChartQA~\cite{masry_chartqa_2022} and Super-CLEVR~\cite{li_super-clevr_2023}.
For ChartQA, we keep the question, age-group labels, legend, and all non-target bars fixed, changing only the first two red Associate-degree bars so that the correct answer moves from ``25 to 34'' to ``35 to 44.''
For Super-CLEVR image 027978, the original scene contains a green truck, a blue bicycle, a yellow bus, and a cyan airplane.
Using the benchmark's pixel-level object mask, we remove only the green truck while preserving the remaining scene content, changing the correct count from four to three.
Figure~\ref{fig:qualitative_visual_gate} presents both counterfactual pairs and
the corresponding model predictions.

In ChartQA, both models produce ``25 to 34'' on the blank image.
After the red-series maximum moves to the second age group, the no-gate variant retains this default answer, whereas Ours updates to ``35 to 44.''
This contrast is consistent with the full model using the changed chart evidence to override a shared default response.
In Super-CLEVR, Ours tracks the $4\!\rightarrow\!3$ object removal and treats the multicolored airplane as one object, while the no-gate variant retains four by splitting the airplane into separate perceived items.
Together, the chart-reading and object-counting interventions complement the quantitative ablation by showing how the visual gate can improve sensitivity to task-relevant visual structure across distinct visual domains.

\textbf{Summary of qualitative observations.}
The comparisons illustrate both a trade-off in reasoning optimization and a complementary visual-grounding behavior.
In the recursive example, the Ours trajectory continues beyond the intermediate milestone and completes the full multi-hop computation, whereas the GRPO trajectory stops early.
In the algebraic example, the Ours trajectory is longer and more granular than the GRPO trajectory but introduces a local arithmetic error.
Although such detailed derivations may be logically well motivated, they can introduce more opportunities for arithmetic error.
In the controlled visual study, Ours follows answer-changing edits in both chart reading and object counting, while the no-gate variant retains its original predictions.
These observations motivate dynamic step-reward scaling for textual reasoning and provide qualitative support for using the visual gate to encourage predictions that track task-relevant image content.

\clearpage
\bibliography{bibliography}

@article{deepseek-ai_deepseek-r1_2025,
  author =        {Guo, Daya and Yang, Dejian and Zhang, Haowei and
                   Song, Junxiao and Wang, Peiyi and Zhu, Qihao and
                   Xu, Runxin and Zhang, Ruoyu and Ma, Shirong and
                   Bi, Xiao and others},
  journal =       {Nature},
  month =         sep,
  number =        {8081},
  pages =         {633--638},
  title =         {{DeepSeek}-{R1} Incentivizes Reasoning in {LLMs}
                   through Reinforcement Learning},
  volume =        {645},
  year =          {2025},
  doi =           {10.1038/s41586-025-09422-z},
  url =           {https://www.nature.com/articles/s41586-025-09422-z},
}

@inproceedings{yue_does_2025,
  author =        {Yue, Yang and Chen, Zhiqi and Lu, Rui and
                   Zhao, Andrew and Wang, Zhaokai and Yue, Yang and
                   Song, Shiji and Huang, Gao},
  month =         nov,
  booktitle =     {Advances in Neural Information Processing Systems},
  pages =         {57654--57689},
  publisher =     {Curran Associates, Inc.},
  title =         {Does {Reinforcement} {Learning} {Really}
                   {Incentivize} {Reasoning} {Capacity} in {LLMs}
                   {Beyond} the {Base} {Model}?},
  volume =        {38},
  year =          {2025},
  url =           {https://proceedings.neurips.cc/paper_files/paper/2025/file/537d5aa768c2d534016a4d06f87bc8fb-Paper-Conference.pdf},
}

@inproceedings{zhang_lessons_2025,
  author =        {Zhang, Zhenru and Zheng, Chujie and Wu, Yangzhen and
                   Zhang, Beichen and Lin, Runji and Yu, Bowen and
                   Liu, Dayiheng and Zhou, Jingren and Lin, Junyang},
  address =       {Vienna, Austria},
  booktitle =     {Findings of the Association for Computational
                   Linguistics: ACL 2025},
  month =         jul,
  pages =         {10495--10516},
  publisher =     {Association for Computational Linguistics},
  title =         {The {Lessons} of {Developing} {Process} {Reward}
                   {Models} in {Mathematical} {Reasoning}},
  year =          {2025},
  doi =           {10.18653/v1/2025.findings-acl.547},
  url =           {https://aclanthology.org/2025.findings-acl.547/},
}

@inproceedings{zhang_linking_2025,
  author =        {Zhang, Zheng and Shan, Ziwei and Song, Kaitao and
                   Li, Yexin and Ren, Kan},
  booktitle =     {The Fourteenth International Conference on Learning
                   Representations},
  month =         apr,
  title =         {Linking {Process} to {Outcome}: {Conditional}
                   {Reward} {Modeling} for {LLM} {Reasoning}},
  year =          {2026},
  url =           {https://openreview.net/forum?id=4DJoBOQNd0},
}

@inproceedings{zhou_reinforcing_2025,
  author =        {Zhou, Xiangxin and Liu, Zichen and Sims, Anya and
                   Wang, Haonan and Pang, Tianyu and Li, Chongxuan and
                   Wang, Liang and Lin, Min and Du, Chao},
  booktitle =     {The Fourteenth International Conference on Learning
                   Representations},
  month =         apr,
  title =         {Reinforcing {General} {Reasoning} without
                   {Verifiers}},
  year =          {2026},
  url =           {https://openreview.net/forum?id=nnwvwge40d},
}

@inproceedings{tang_beyond_2025,
  author =        {Tang, Yunhao and Wang, Sid and Madaan, Lovish and
                   Munos, Rémi},
  month =         may,
  booktitle =     {Advances in Neural Information Processing Systems},
  pages =         {74421--74448},
  publisher =     {Curran Associates, Inc.},
  title =         {Beyond {Verifiable} {Rewards}: {Scaling}
                   {Reinforcement} {Learning} for {Language} {Models} to
                   {Unverifiable} {Data}},
  volume =        {38},
  year =          {2025},
  url =           {https://proceedings.neurips.cc/paper_files/paper/2025/file/6bd67a424dc59481e1e5a5061ffc8dfe-Paper-Conference.pdf},
}

@article{shao_deepseekmath_2024,
  author =        {Shao, Zhihong and Wang, Peiyi and Zhu, Qihao and
                   Xu, Runxin and Song, Junxiao and Bi, Xiao and
                   Zhang, Haowei and Zhang, Mingchuan and Li, Y. K. and
                   Wu, Y. and others},
  month =         apr,
  journal =       {arXiv preprint arXiv:2402.03300},
  title =         {{DeepSeekMath}: {Pushing} the {Limits} of
                   {Mathematical} {Reasoning} in {Open} {Language}
                   {Models}},
  year =          {2024},
  doi =           {10.48550/arXiv.2402.03300},
  url =           {http://arxiv.org/abs/2402.03300},
}

@inproceedings{yu_dapo_2025,
  author =        {Yu, Qiying and Zhang, Zheng and Zhu, Ruofei and
                   Yuan, Yufeng and Zuo, Xiaochen and Yue, Yu and
                   Dai, Weinan and Fan, Tiantian and Liu, Gaohong and
                   Liu, Juncai and others},
  booktitle =     {Advances in Neural Information Processing Systems},
  month =         may,
  pages =         {113222--113244},
  publisher =     {Curran Associates, Inc.},
  title =         {{DAPO}: {An} {Open}-{Source} {LLM} {Reinforcement}
                   {Learning} {System} at {Scale}},
  volume =        {38},
  year =          {2025},
  url =           {https://proceedings.neurips.cc/paper_files/paper/2025/file/a4277440d50f1f15d2cb4c14f7e0c0d2-Paper-Conference.pdf},
}

@article{zheng_group_2025,
  author =        {Zheng, Chujie and Liu, Shixuan and Li, Mingze and
                   Chen, Xiong-Hui and Yu, Bowen and Gao, Chang and
                   Dang, Kai and Liu, Yuqiong and Men, Rui and Yang, An and
                   others},
  month =         jul,
  journal =       {arXiv preprint arXiv:2507.18071},
  title =         {Group {Sequence} {Policy} {Optimization}},
  year =          {2025},
  doi =           {10.48550/arXiv.2507.18071},
  url =           {http://arxiv.org/abs/2507.18071},
}

@article{liu_gdpo_2026,
  author =        {Liu, Shih-Yang and Dong, Xin and Lu, Ximing and
                   Diao, Shizhe and Belcak, Peter and Liu, Mingjie and
                   Chen, Min-Hung and Yin, Hongxu and
                   Wang, Yu-Chiang Frank and Cheng, Kwang-Ting and others},
  month =         jan,
  journal =       {arXiv preprint arXiv:2601.05242},
  title =         {{GDPO}: {Group} reward-{Decoupled} {Normalization}
                   {Policy} {Optimization} for {Multi}-reward {RL}
                   {Optimization}},
  year =          {2026},
  doi =           {10.48550/arXiv.2601.05242},
  url =           {http://arxiv.org/abs/2601.05242},
}

@inproceedings{zhang_r1-vl_2025,
  author =        {Zhang, Jingyi and Huang, Jiaxing and Yao, Huanjin and
                   Liu, Shunyu and Zhang, Xikun and Lu, Shijian and
                   Tao, Dacheng},
  booktitle =     {Proceedings of the IEEE/CVF International Conference
                   on Computer Vision (ICCV)},
  month =         oct,
  pages =         {1859--1869},
  title =         {R1-{VL}: {Learning} to {Reason} with {Multimodal}
                   {Large} {Language} {Models} via {Step}-wise {Group}
                   {Relative} {Policy} {Optimization}},
  year =          {2025},
  url =           {https://openaccess.thecvf.com/content/ICCV2025/html/Zhang_R1-VL_Learning_to_Reason_with_Multimodal_Large_Language_Models_via_ICCV_2025_paper.html},
}

@article{chen_language_2024,
  author =        {Chen, Haolin and Feng, Yihao and Liu, Zuxin and
                   Yao, Weiran and Prabhakar, Akshara and Heinecke, Shelby and
                   Ho, Ricky and Mui, Phil and Savarese, Silvio and
                   Xiong, Caiming and others},
  month =         nov,
  journal =       {arXiv preprint arXiv:2411.04282},
  title =         {Language {Models} are {Hidden} {Reasoners}:
                   {Unlocking} {Latent} {Reasoning} {Capabilities} via
                   {Self}-{Rewarding}},
  year =          {2024},
  doi =           {10.48550/arXiv.2411.04282},
  url =           {http://arxiv.org/abs/2411.04282},
}

@inproceedings{zelikman_star_2022,
  author =        {Zelikman, Eric and Wu, Yuhuai and Mu, Jesse and
                   Goodman, Noah D.},
  month =         may,
  booktitle =     {Advances in Neural Information Processing Systems},
  pages =         {15476--15488},
  publisher =     {Curran Associates, Inc.},
  title =         {{STaR}: {Bootstrapping} {Reasoning} {With}
                   {Reasoning}},
  volume =        {35},
  year =          {2022},
  doi =           {10.52202/068431-1126},
  url =           {https://proceedings.neurips.cc/paper_files/paper/2022/hash/639a9a172c044fbb64175b5fad42e9a5-Abstract-Conference.html},
}

@article{gulcehre_reinforced_2023,
  author =        {Gulcehre, Caglar and Paine, Tom Le and
                   Srinivasan, Srivatsan and Konyushkova, Ksenia and
                   Weerts, Lotte and Sharma, Abhishek and Siddhant, Aditya and
                   Ahern, Alex and Wang, Miaosen and Gu, Chenjie and others},
  month =         aug,
  journal =       {arXiv preprint arXiv:2308.08998},
  title =         {Reinforced {Self}-{Training} ({ReST}) for {Language}
                   {Modeling}},
  year =          {2023},
  doi =           {10.48550/arXiv.2308.08998},
  url =           {http://arxiv.org/abs/2308.08998},
}

@article{cobbe_training_2021,
  author =        {Cobbe, Karl and Kosaraju, Vineet and Bavarian, Mohammad and
                   Chen, Mark and Jun, Heewoo and Kaiser, Lukasz and
                   Plappert, Matthias and Tworek, Jerry and Hilton, Jacob and
                   Nakano, Reiichiro and others},
  month =         nov,
  journal =       {arXiv preprint arXiv:2110.14168},
  title =         {Training {Verifiers} to {Solve} {Math} {Word}
                   {Problems}},
  year =          {2021},
  doi =           {10.48550/arXiv.2110.14168},
  url =           {http://arxiv.org/abs/2110.14168},
}

@inproceedings{hendrycks_measuring_2021,
  author =        {Hendrycks, Dan and Burns, Collin and Kadavath, Saurav and
                   Arora, Akul and Basart, Steven and Tang, Eric and
                   Song, Dawn and Steinhardt, Jacob},
  month =         nov,
  booktitle =     {Proceedings of the Neural Information Processing
                   Systems Track on Datasets and Benchmarks},
  publisher =     {Curran Associates, Inc.},
  title =         {Measuring {Mathematical} {Problem} {Solving} {With}
                   the {MATH} {Dataset}},
  volume =        {1},
  year =          {2021},
  url =           {https://datasets-benchmarks-proceedings.neurips.cc/paper_files/paper/2021/hash/be83ab3ecd0db773eb2dc1b0a17836a1-Abstract-round2.html},
}

@article{liu_matheval_2025,
  author =        {Liu, Tianqiao and Chen, Zui and Fang, Zhensheng and
                   Luo, Weiqi and Tian, Mi and Liu, Zitao},
  journal =       {Frontiers of Digital Education},
  month =         jun,
  number =        {2},
  pages =         {16},
  title =         {{MathEval}: {A} {Comprehensive} {Benchmark} for
                   {Evaluating} {Large} {Language} {Models} on
                   {Mathematical} {Reasoning} {Capabilities}},
  volume =        {2},
  year =          {2025},
  doi =           {10.1007/s44366-025-0053-z},
  issn =          {2097-3918, 2097-3926},
  url =           {https://link.springer.com/10.1007/s44366-025-0053-z},
}

@inproceedings{liu_cmm-math_2025,
  address =       {Dublin Ireland},
  author =        {Liu, Wentao and Pan, Qianjun and Zhang, Yi and
                   Liu, Zhuo and Wu, Ji and Zhou, Jie and Zhou, Aimin and
                   Chen, Qin and Jiang, Bo and He, Liang},
  booktitle =     {Proceedings of the 33rd {ACM} {International}
                   {Conference} on {Multimedia}},
  month =         oct,
  pages =         {12585--12591},
  publisher =     {ACM},
  title =         {{CMM}-{Math}: {A} {Chinese} {Multimodal} {Math}
                   {Dataset} {To} {Evaluate} and {Enhance} the
                   {Mathematics} {Reasoning} of {Large} {Multimodal}
                   {Models}},
  year =          {2025},
  doi =           {10.1145/3746027.3758193},
  isbn =          {979-8-4007-2035-2},
  url =           {https://dl.acm.org/doi/10.1145/3746027.3758193},
}

@inproceedings{masry_chartqa_2022,
  address =       {Dublin, Ireland},
  author =        {Masry, Ahmed and Long, Do and Tan, Jia Qing and
                   Joty, Shafiq and Hoque, Enamul},
  booktitle =     {Findings of the {Association} for {Computational}
                   {Linguistics}: {ACL} 2022},
  pages =         {2263--2279},
  publisher =     {Association for Computational Linguistics},
  title =         {{ChartQA}: {A} {Benchmark} for {Question} {Answering}
                   about {Charts} with {Visual} and {Logical}
                   {Reasoning}},
  year =          {2022},
  doi =           {10.18653/v1/2022.findings-acl.177},
  url =           {https://aclanthology.org/2022.findings-acl.177},
}

@inproceedings{li_super-clevr_2023,
  address =       {Vancouver, BC, Canada},
  author =        {Li, Zhuowan and Wong, Xingrui and
                   Stengel-Eskin, Elias and Kortylewski, Adam and
                   Ma, Wufei and Van Durme, Benjamin and Yuille, Alan},
  booktitle =     {2023 {IEEE}/{CVF} {Conference} on {Computer} {Vision}
                   and {Pattern} {Recognition} ({CVPR})},
  month =         jun,
  pages =         {14963--14973},
  publisher =     {IEEE},
  title =         {Super-{CLEVR}: {A} {Virtual} {Benchmark} to
                   {Diagnose} {Domain} {Robustness} in {Visual}
                   {Reasoning}},
  year =          {2023},
  doi =           {10.1109/CVPR52729.2023.01437},
  isbn =          {979-8-3503-0129-8},
  url =           {https://ieeexplore.ieee.org/document/10205295/},
}

@article{chen_g1_2025,
  author =        {Chen, Liang and Gao, Hongcheng and Liu, Tianyu and
                   Huang, Zhiqi and Sung, Flood and Zhou, Xinyu and
                   Wu, Yuxin and Chang, Baobao},
  month =         may,
  journal =       {arXiv preprint arXiv:2505.13426},
  title =         {G1: {Bootstrapping} {Perception} and {Reasoning}
                   {Abilities} of {Vision}-{Language} {Model} via
                   {Reinforcement} {Learning}},
  year =          {2025},
  doi =           {10.48550/arXiv.2505.13426},
  url =           {http://arxiv.org/abs/2505.13426},
}

@article{wang2024mmlupro,
  author =        {Wang, Yubo and Ma, Xueguang and Zhang, Ge and
                   Ni, Yuansheng and Chandra, Abhranil and Guo, Shiguang and
                   Ren, Weiming and Arulraj, Aaran and He, Xuan and
                   Jiang, Ziyan and others},
  journal =       {Advances in Neural Information Processing Systems},
  pages =         {95266--95290},
  title =         {MMLU-Pro: A more robust and challenging multi-task
                   language understanding benchmark},
  volume =        {37},
  year =          {2024},
}

@inproceedings{talmor2019commonsenseqa,
  author =        {Talmor, Alon and Herzig, Jonathan and
                   Lourie, Nicholas and Berant, Jonathan},
  booktitle =     {Proceedings of the 2019 Conference of the North
                   American Chapter of the Association for Computational
                   Linguistics: Human Language Technologies, Volume 1
                   (Long and Short Papers)},
  pages =         {4149--4158},
  title =         {Commonsenseqa: A question answering challenge
                   targeting commonsense knowledge},
  year =          {2019},
}

@article{meng2025mmeureka,
  author =        {Meng, Fanqing and Du, Lingxiao and Liu, Zongkai and
                   Zhou, Zhixiang and Lu, Quanfeng and Fu, Daocheng and
                   Han, Tiancheng and Shi, Botian and Wang, Wenhai and
                   He, Junjun and others},
  journal =       {arXiv preprint arXiv:2503.07365},
  title =         {MM-Eureka: Exploring the frontiers of multimodal
                   reasoning with rule-based reinforcement learning},
  year =          {2025},
}

@inproceedings{lu2023mathvista,
  author =        {Lu, Pan and Bansal, Hritik and Xia, Tony and
                   Liu, Jiacheng and Li, Chunyuan and
                   Hajishirzi, Hannaneh and Cheng, Hao and
                   Chang, Kai-Wei and Galley, Michel and Gao, Jianfeng},
  booktitle =     {The Twelfth International Conference on Learning
                   Representations},
  title =         {{MathVista}: Evaluating Mathematical Reasoning of
                   Foundation Models in Visual Contexts},
  url =           {https://openreview.net/forum?id=KUNzEQMWU7},
  year =          {2024},
}

@article{chen2024we,
  author =        {Chen, Lin and Li, Jinsong and Dong, Xiaoyi and
                   Zhang, Pan and Zang, Yuhang and Chen, Zehui and
                   Duan, Haodong and Wang, Jiaqi and Qiao, Yu and
                   Lin, Dahua and others},
  journal =       {Advances in Neural Information Processing Systems},
  pages =         {27056--27087},
  title =         {Are we on the right way for evaluating large
                   vision-language models?},
  volume =        {37},
  year =          {2024},
}

@inproceedings{zhang2024mathverse,
  author =        {Zhang, Renrui and Jiang, Dongzhi and Zhang, Yichi and
                   Lin, Haokun and Guo, Ziyu and Qiu, Pengshuo and
                   Zhou, Aojun and Lu, Pan and Chang, Kai-Wei and Qiao, Yu and
                   others},
  booktitle =     {European Conference on Computer Vision},
  pages =         {169--186},
  title =         {Mathverse: Does your multi-modal LLM truly see the
                   diagrams in visual math problems?},
  year =          {2024},
}

@inproceedings{lightman_verify_2024,
  author =        {Lightman, Hunter and Kosaraju, Vineet and Burda, Yuri and
                   Edwards, Harrison and Baker, Bowen and Lee, Teddy and
                   Leike, Jan and Schulman, John and Sutskever, Ilya and
                   Cobbe, Karl},
  booktitle =     {The Twelfth International Conference on Learning
                   Representations},
  title =         {Let's Verify Step by Step},
  url =           {https://openreview.net/forum?id=v8L0pN6EOi},
  year =          {2024},
}

@inproceedings{wang_math_shepherd_2024,
  address =       {Bangkok, Thailand},
  author =        {Wang, Peiyi and Li, Lei and Shao, Zhihong and Xu, Runxin and
                   Dai, Damai and Li, Yifei and Chen, Deli and Wu, Yu and
                   Sui, Zhifang},
  booktitle =     {Proceedings of the 62nd Annual Meeting of the Association
                   for Computational Linguistics (Volume 1: Long Papers)},
  doi =           {10.18653/v1/2024.acl-long.510},
  month =         aug,
  pages =         {9426--9439},
  publisher =     {Association for Computational Linguistics},
  title =         {{Math-Shepherd}: Verify and Reinforce {LLMs} Step-by-step
                   without Human Annotations},
  url =           {https://aclanthology.org/2024.acl-long.510/},
  year =          {2024},
}

@inproceedings{arjona_medina_rudder_2019,
  author =        {Arjona-Medina, Jose A. and Gillhofer, Michael and
                   Widrich, Michael and Unterthiner, Thomas and
                   Brandstetter, Johannes and Hochreiter, Sepp},
  booktitle =     {Advances in Neural Information Processing Systems},
  publisher =     {Curran Associates, Inc.},
  title =         {{RUDDER}: Return Decomposition for Delayed Rewards},
  url =           {https://proceedings.neurips.cc/paper_files/paper/2019/file/16105fb9cc614fc29e1bda00dab60d41-Paper.pdf},
  volume =        {32},
  year =          {2019},
}

@inproceedings{niu_counterfactual_2021,
  author =        {Niu, Yulei and Tang, Kaihua and Zhang, Hanwang and
                   Lu, Zhiwu and Hua, Xian-Sheng and Wen, Ji-Rong},
  booktitle =     {Proceedings of the IEEE/CVF Conference on Computer Vision
                   and Pattern Recognition (CVPR)},
  month =         jun,
  pages =         {12700--12710},
  title =         {Counterfactual {VQA}: A Cause-Effect Look at Language Bias},
  url =           {https://openaccess.thecvf.com/content/CVPR2021/html/Niu_Counterfactual_VQA_A_Cause-Effect_Look_at_Language_Bias_CVPR_2021_paper.html},
  year =          {2021},
}

@article{qwen25_2024,
  author =        {{Qwen Team}},
  doi =           {10.48550/arXiv.2412.15115},
  journal =       {arXiv preprint arXiv:2412.15115},
  title =         {{Qwen2.5} Technical Report},
  url =           {https://arxiv.org/abs/2412.15115},
  year =          {2024},
}

@article{bai_qwen25vl_2025,
  author =        {Bai, Shuai and Chen, Keqin and Liu, Xuejing and
                   Wang, Jialin and Ge, Wenbin and Song, Sibo and Dang, Kai and
                   Wang, Peng and Wang, Shijie and Tang, Jun and others},
  doi =           {10.48550/arXiv.2502.13923},
  journal =       {arXiv preprint arXiv:2502.13923},
  title =         {{Qwen2.5-VL} Technical Report},
  url =           {https://arxiv.org/abs/2502.13923},
  year =          {2025},
}

@article{schulman_ppo_2017,
  author =        {Schulman, John and Wolski, Filip and Dhariwal, Prafulla and
                   Radford, Alec and Klimov, Oleg},
  doi =           {10.48550/arXiv.1707.06347},
  journal =       {arXiv preprint arXiv:1707.06347},
  title =         {Proximal Policy Optimization Algorithms},
  url =           {https://arxiv.org/abs/1707.06347},
  year =          {2017},
}

@inproceedings{rajbhandari_zero_2020,
  author =        {Rajbhandari, Samyam and Rasley, Jeff and Ruwase, Olatunji and
                   He, Yuxiong},
  booktitle =     {Proceedings of the International Conference for High
                   Performance Computing, Networking, Storage and Analysis},
  doi =           {10.1109/SC41405.2020.00024},
  pages =         {1--16},
  title =         {{ZeRO}: Memory Optimizations Toward Training Trillion
                   Parameter Models},
  url =           {https://sc20.supercomputing.org/proceedings/tech_paper/tech_paper_pages/pap379.html},
  year =          {2020},
}

@inproceedings{johnson_clevr_2017,
  author =        {Johnson, Justin and Hariharan, Bharath and
                   van der Maaten, Laurens and Fei-Fei, Li and
                   Zitnick, C. Lawrence and Girshick, Ross},
  booktitle =     {Proceedings of the IEEE Conference on Computer Vision and
                   Pattern Recognition (CVPR)},
  month =         jul,
  pages =         {2901--2910},
  title =         {{CLEVR}: A Diagnostic Dataset for Compositional Language
                   and Elementary Visual Reasoning},
  url =           {https://openaccess.thecvf.com/content_cvpr_2017/html/Johnson_CLEVR_A_Diagnostic_CVPR_2017_paper.html},
  year =          {2017},
}

@article{uesato_process_outcome_2022,
  author =        {Uesato, Jonathan and Kushman, Nate and Kumar, Ramana and
                   Song, Francis and Siegel, Noah and Wang, Lisa and
                   Creswell, Antonia and Irving, Geoffrey and Higgins, Irina},
  doi =           {10.48550/arXiv.2211.14275},
  journal =       {arXiv preprint arXiv:2211.14275},
  title =         {Solving Math Word Problems with Process- and Outcome-Based
                   Feedback},
  url =           {https://arxiv.org/abs/2211.14275},
  year =          {2022},
}

@article{cui_prime_2026,
  author =        {Cui, Ganqu and Yuan, Lifan and Wang, Zefan and Wang, Hanbin and
                   Zhang, Yuchen and Chen, Jiacheng and Li, Wendi and
                   He, Bingxiang and Fan, Yuchen and Yu, Tianyu and others},
  journal =       {Transactions on Machine Learning Research},
  title =         {Process Reinforcement through Implicit Rewards},
  url =           {https://openreview.net/forum?id=9SkkifLopZ},
  year =          {2026},
}

@inproceedings{kazemnejad_vineppo_2025,
  author =        {Kazemnejad, Amirhossein and Aghajohari, Milad and
                   Portelance, Eva and Sordoni, Alessandro and Reddy, Siva and
                   Courville, Aaron and Le Roux, Nicolas},
  booktitle =     {Proceedings of the 42nd International Conference on Machine
                   Learning},
  pages =         {29557--29590},
  publisher =     {PMLR},
  series =        {Proceedings of Machine Learning Research},
  title =         {{VinePPO}: Refining Credit Assignment in {RL} Training of
                   {LLMs}},
  url =           {https://proceedings.mlr.press/v267/kazemnejad25a.html},
  volume =        {267},
  year =          {2025},
}

@article{patel2021nlp,
  author =        {Patel, Arkil and Bhattamishra, Satwik and Goyal, Navin},
  journal =       {arXiv preprint arXiv:2103.07191},
  title =         {Are {NLP} Models Really Able to Solve Simple Math Word
                   Problems?},
  year =          {2021},
}

\end{document}